%% file: main.tex
\documentclass[11pt]{article}

\usepackage[final]{acl}   %

\usepackage{times}
\usepackage{latexsym}
\usepackage[T1]{fontenc}
\usepackage[utf8]{inputenc}
\usepackage{microtype}
\usepackage{graphicx}
\usepackage{booktabs}
\usepackage{amsmath}
\usepackage{xcolor}
\usepackage{multirow}
\usepackage{listings}

\input{numbers}   %
\newcommand{\pp}{\,pp}   %
\makeatletter\g@addto@macro{\UrlBreaks}{\do\-}\makeatother   %
\makeatletter   %
\newcommand{\coderepo}{\url{https://github.com/Yangxinyee/steering-out-of-the-loop}}
\newcommand{\genrepo}{\url{https://huggingface.co/datasets/yangxinye/steering-out-of-the-loop-generations}}
\makeatother

\title{Steering Out of the Loop: Does Weight-Space Programming Add Reasoning Ability?}

\author{%
  Xinye Yang \\
  {\small Department of Radiology} \\
  {\small University of Colorado Anschutz School of Medicine} \\
  {\small Aurora, CO, USA} \\
  {\small\texttt{xinye\_yang@alumni.brown.edu}} \\
  \And
  Yuli Wang \\
  {\small Department of Radiology} \\
  {\small Johns Hopkins University School of Medicine} \\
  {\small Baltimore, MD, USA} \\
  {\small\texttt{ywang687@jhmi.edu}} \\
  \AND
  Harrison Bai\thanks{Corresponding author.} \\
  {\small Department of Radiology} \\
  {\small University of Colorado Anschutz School of Medicine} \\
  {\small Aurora, CO, USA} \\
  {\small\texttt{harrison.bai@cuanschutz.edu}} \\
}
\AtBeginDocument{\hypersetup{pdftitle={Steering Out of the Loop: Does Weight-Space Programming Add Reasoning Ability?},
  pdfauthor={Xinye Yang, Yuli Wang, Harrison Bai}}}

\begin{document}
\maketitle

\begin{abstract}
Accuracy gains in reasoning can come from releasing ability a model already has or from adding problems it can newly
solve; for the headline edit of weight-space programming, we find mostly release. For looping reasoning models, we
introduce the beyond-reach test: it estimates the unedited model's empirical reach under eight decoding fixes and reads
correct answers beyond that reach against decoding-only references, fine-tunes on the model's own answers and a
distilled positive control. In the audited edit, an AI agent changed \wAuthorsEditShare\% of Qwen3.5-0.8B's weights,
raising MATH-500 accuracy by a reported \wAuthorsGainMath{} points. Disabling thinking alone recovers \wOffShareMath\% of
the gain we reproduce, and nearly all of the edit's correct answers on always-looping problems lie within the decoding
reach, as do the controls'. Matched on finished samples, the unedited model with thinking off solves more problems beyond
reach than the edit on MATH-500 and GSM8K. The test detects distillation on GSM8K but only bounds additions on MATH-500
and MMLU-Pro math. What the edit keeps is single-sample reliability, which voting over thinking-off samples at no more
tokens matches or exceeds on MATH-500 and GSM8K. Before crediting a weight edit with new reasoning ability, count its
correct answers beyond the decoding reach.
\end{abstract}

\begin{figure*}[t]
\centering
\includegraphics[width=\textwidth]{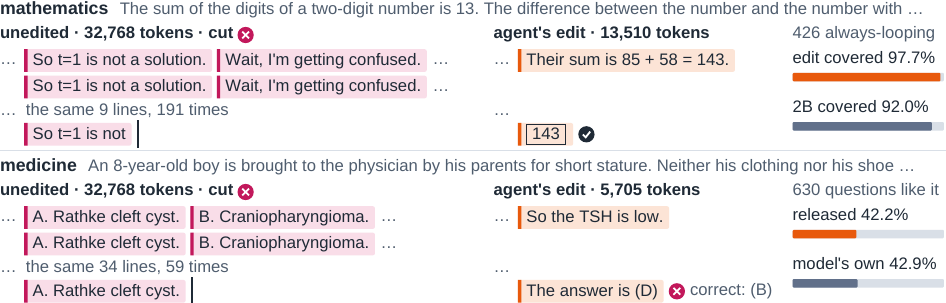}
\caption{Top: a MATH-500 loop the agent's edit ends, drawn at random; right, the share of the edit's and Qwen3.5-2B's
correct answers on always-looping problems that lie within the decoding reach. Bottom: a MedQA question; right, released
against own accuracy on the same questions.}
\label{fig:teaser}
\end{figure*}

\input{tables/headline}

\section{Introduction}
\label{sec:intro}

When an intervention raises a model's accuracy, it can do so in two ways. It can \emph{release} ability: the model
solves, more reliably, problems it could already solve. Or it can \emph{add} ability: the model solves problems beyond
its empirical reach. Accuracy measures how often a model succeeds, not which problems it can solve, so it cannot tell
the two apart. The difference is large for small reasoning models, which often fail by not finishing: Qwen3.5-0.8B
runs to its 32{,}768-token limit on \wBaseCutMath\% of MATH-500 answers, repeating itself until it is cut off (Figure~\ref{fig:teaser}). Any
change that ends such loops raises accuracy, whether or not it adds ability.

Weight-space programming makes the question concrete. An AI agent with a forward-pass tool and no gradients edited
Qwen3.5-0.8B's weights directly, changing \wAuthorsEditShare\% of them (64 columns in each of two MLP output matrices),
and raised MATH-500 accuracy from \wAuthorsBaseMath\% to \wAuthorsEditMath\% \citep{takehi2026sparks}. The agent was
asked to make the model ``better at mathematics'', with repairing repetition in scope (Appendix~\ref{app:task}). A gain
this large from an edit this small is exactly where release and addition must be told apart, and no one has measured
which of the two it is on a model that loops.

We introduce the \emph{beyond-reach test} for looping reasoning models (\S\ref{sec:method}).
The \emph{decoding reach} is the set of problems the unedited model solves when sampled under eight decoding fixes, each
of which changes only how the model decodes. An intervention that only releases ability answers correctly within the
model's full reach; a finite decoding reach misses some of those problems, so the test reads the count beyond it against
two references that only decode and so release by construction, the unedited model sampled again and the unedited model
with thinking off, each matched to the finished samples it has, and against fine-tunes on the model's own answers. A
0.8B distilled from a 9B model is the positive control: it shows where the test detects an addition and where it can
only bound one.

Applied to the agent's edit, the beyond-reach test answers five questions (Table~\ref{tab:headline}). \textbf{RQ1:}
thinking off, with no weight change, recovers \wOffShareMath\% of the MATH-500 gain we reproduce (\wBaseMath\% to
\wEditMath\%). \textbf{RQ2:} where the model always loops, \wCovMath\% and \wCovGsm\% of the edit's correct MATH-500
and GSM8K answers lie within the decoding reach, as do those of the controls. \textbf{RQ3:} matched on finished samples,
the unedited model with thinking off solves more problems beyond reach than the edit on MATH-500 (\wRaMathEditTo{}
\wRaMathEditToCI) and GSM8K, and on MMLU-Pro math every count bounds the edit. The positive control is detected on GSM8K;
on MATH-500 it does not exceed thinking off, so there the test only bounds. \textbf{RQ4:} on whole test sets one strict
sample of the edit is right more often than one sample of the best decoding fix, by \wResidLo--\wResidHi{} points.  The lead is per sample, not per token: at no more tokens, voting over thinking-off samples
beats it on MATH-500 and ties it on GSM8K. \textbf{RQ5:} in medicine the edit's gain over a temperature change is
\wMedEditMinusTone{} \wMedEditMinusToneCI{} points, and on a second model the textbook edit's correct answers all lie
within reach.

Our contributions are:
\begin{itemize}
  \setlength\itemsep{0pt}
  \item The beyond-reach test for looping reasoning models: a decoding reach pooled from eight decoding fixes, and a
  count beyond it read against references that only decode, matched on finished samples, against fine-tunes on the
  model's own answers and against a distilled positive control that shows where the test detects an addition (GSM8K)
  and where it only bounds one (MATH-500). Its controls show that fine-tuning on answers within reach can itself solve more beyond reach than both
  decoding-only references (MMLU-Pro math), and that an accuracy criterion credits finishing more answers.
  \item An audit of the headline edit of weight-space programming: thinking off recovers most of its gain, and matched decoding solves more problems beyond reach than it does on MATH-500
and GSM8K. What it
  keeps is single-sample reliability on whole test sets, measured per strict sample, on the samples both settings finish
  and against voting at no more tokens.
  
\end{itemize}
{\raggedright Prompts, per-problem outputs and analysis code are released (\coderepo), the raw generations
on Hugging Face (\genrepo).\par}

\begin{figure*}[t]
\centering
\includegraphics[width=\textwidth]{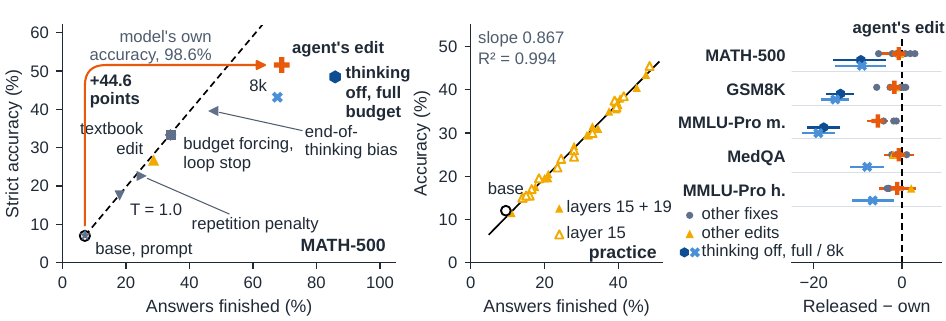}
\caption{Left, strict accuracy against answers finished (arrow: the edit's strict gain; dashed slope: own accuracy);
middle, 32 textbook edits on 200 practice problems; right, released minus own accuracy on the same problems (points,
95\% intervals). No fix beats the model's own.}
\label{fig:completion}
\end{figure*}

\section{Related Work}
\label{sec:related}

\paragraph{Released or added ability.} Using pass@$k$ to ask whether training releases or adds ability comes from work
on reinforcement learning with verifiable rewards: base models reach at large $k$ what RL-trained ones reach at small
$k$ \citep{yue2025rl}, and RL may deploy existing capabilities rather than add new ones \citep{venhoff2026base}.
Coverage grows with repeated sampling \citep{brown2024monkeys}, and test-time compute can stand in for parameters
\citep{snell2024scaling}. We differ in four ways: our models loop, so on many problems their own decoding never finishes and the reach must come from decoding fixes; the reach pools eight such fixes; the count beyond it is read against references that only decode, matched on finished samples, and against a positive
control; and a flag judges one model alone.

\paragraph{Steering and editing reasoning models.}
Activation steering and weight editing shorten or recalibrate reasoning: SEAL \citep{chen2025seal}, Manifold Steering
\citep{huang2025manifold}, ThinkEdit \citep{sun2025thinkedit} and Steer2Edit \citep{sun2026steer2edit} report aggregate
accuracy and tokens; we ask whether an intervention's answers are released or added.

\paragraph{Repetition loops.}
Repetition is a long-known failure of text generation \citep{holtzman2020degeneration,welleck2020unlikelihood,xu2022loop}.
Temperature reduces looping in reasoning models without fixing its cause \citep{pipis2026wait}, and hidden states
detect it early \citep{xie2025wordsalad,duan2026circular}. Concurrent work steers pruned reasoning models toward
termination and counts generations that become correct \citep{nguyen2026tgas}, and single-neuron edits suppress
repetition, which cannot supply a missing fact \citep{lazaridis2026oneneuron}; we measure such gains against tuned
decoding fixes.

\paragraph{Weight-space programming.} In weight-space programming a general-purpose model reads and writes a network's
weights directly; the run we audit is logged in full \citep{takehi2026sparks}. Follow-up work drives weight edits by
experimental feedback \citep{li2026weightloop}, audits small clinical networks \citep{nusrat2026clinical,liu2026computation}
and studies the reading side \citep{ahmed2026memorization,liu2026computation}; none tests what the reasoning edit adds.
Password-locked models draw the line between eliciting and teaching by construction \citep{greenblatt2024password}.

\section{Measuring Added Ability under Decoding Fixes}
\label{sec:method}

\paragraph{Setting.} Qwen3.5-0.8B \citep{qwen2026qwen35} is the main model; unedited, it is also called the \emph{base model}. We follow the evaluation protocol of the
edit's authors \citep{takehi2026sparks}: thinking on, temperature 0.6 (the model card's is 1.0), top-$p$ 0.95, top-$k$
20, presence penalty 1.5 and a budget of 32{,}768 generated tokens, served with vLLM \citep{kwon2023vllm}. An answer
that reaches the budget is \emph{cut}; otherwise it is \emph{finished}. Each sample's seed is set by its problem and
index, the same in every setting. Test sets are MATH-500 \citep{hendrycks2021math,lightman2024verify}, GSM8K
\citep{cobbe2021gsm8k}, the mathematics (1{,}351 questions) and health (818) categories of MMLU-Pro
\citep{wang2024mmlupro} and MedQA \citep{jin2021medqa} (1{,}273), four samples per problem. Every hyperparameter is set
on practice problems from the training splits (Appendix~\ref{app:protocol}). In layers 15 and 19 the agent replaced the
MLP output matrix $W$ by $W - c\,u\,(u^\top W_{:,J})$ on a set $J$ of 64 columns, with $c = 6$ and a unit direction $u$
per layer computed from repetitive and normal contexts. The \emph{textbook edit} takes the same form with $u$ the
difference of mean residuals between looping and productive stretches of the model's own answers
\citep{rimsky2024caa}, and $J$ the 64 neurons that write most along it.

\paragraph{Decoding fixes.} Eight fixes change only decoding. \emph{Thinking off} \citep{ma2025nothinking} disables
the reasoning mode (temperature 0.7, top-$p$ 0.8), at the full budget and at 8{,}192 tokens. \emph{$T = 1.0$} samples
at the model card's temperature. \emph{Budget forcing} \citep{muennighoff2025s1} ends thinking at 8{,}192 tokens and has
the model answer within \wForcedCap{} more tokens; \emph{loop stop} does so when a loop detector fires (a block of
1--200 words repeated three times in a row, the authors' rule, or mostly repeated four-word phrases in the last 200
words). The others are an end-of-thinking bias (a logit bonus for the token that closes thinking), a repetition penalty
\citep{keskar2019ctrl} and a prompt asking the model to stop repeating.

\paragraph{Decoding reach.} A problem is \emph{always-looping} when all four samples of the unedited model under the
protocol are cut. On these problems the unedited model answers only through a decoding fix. Its \emph{decoding reach}
at $k$ is the set of always-looping problems that at least one of $k$ of its samples under the eight fixes finishes
correctly; a cut sample never counts. Unless stated, $k = \wRulerSamples$, four samples under each fix. At $k = 16$ and
$4$ the reach is the expected one of $k$ samples drawn at random from these \wRulerSamples{} (the unbiased pass@$k$
estimator \citep{chen2021codex}); the thinking-on reach pools the six fixes that keep thinking on (24 samples). A correct
answer lies \emph{within reach} if its problem does, otherwise \emph{beyond reach} (terms: Table~\ref{tab:terms}).

\begin{figure*}[t]
\centering
\includegraphics[width=\textwidth]{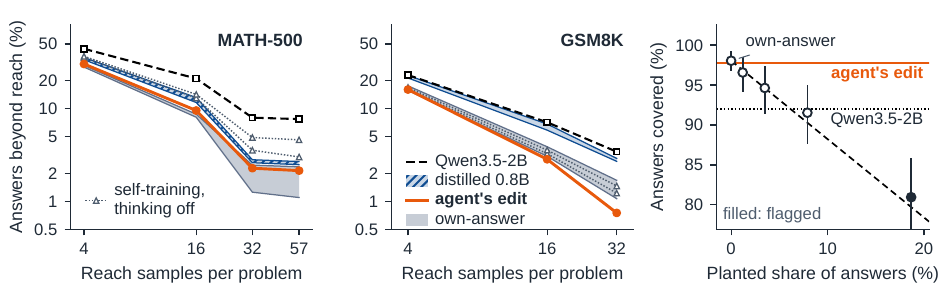}
\caption{Left, middle: correct answers beyond the decoding reach of $k$ samples on always-looping problems (log scale);
bands span the six own-answer runs (grey) and three distilled runs (hatched); dotted: self-training. Right: planted
fine-tunes (MATH-500); filled: flagged.}
\label{fig:reach}
\end{figure*}

\paragraph{Beyond reach, per answer and per problem.} Per answer, a model's \emph{beyond-reach share} is the share of its
correct answers on always-looping problems (four samples each) that lie beyond reach; its complement is the model's
\emph{coverage}. Per problem, with 32 samples of a model's own, its \emph{own-32 count} is the number of problems
beyond reach that some finished sample of it solves. The share asks where a model's answers fall; the count asks how
many problems outside the reach the model solves given as many samples as the reach has. On MMLU-Pro math, where uniform guesses among about ten options solve questions by chance, both are chance-corrected:
the share as $(c - g)/(1 - g)$, with $g$ the share uniform guesses would reach from the same finished samples, and the
count by subtracting, per question, the chance that such guesses solve it.

\paragraph{References and controls.}  Two \emph{decoding-only references} release by construction: the unedited model
\emph{sampled again} (four new samples under each fix) and the unedited model with \emph{thinking off} at the full
budget (32 further samples). \emph{Own-answer fine-tunes} learn only answers within reach: the 0.8B fine-tuned on its own
correct answers under the end-of-thinking bias (\emph{bias answers}), \wSeedsNMath{} seeds and \wAlignedNGsm{} runs on
the distilled model's problems (Appendix~\ref{app:sft}). The \emph{distilled 0.8B}, fine-tuned on Qwen3.5-9B's correct
solutions to \wSftExamples{} training problems (\wDistNGsm{} seeds), is the positive control. Qwen3.5-2B and
\emph{self-training} on the model's own thinking-off answers, one round of STaR \citep{zelikman2022star}, are further
references.

\input{tables/main}

\paragraph{Reading.} The beyond-reach test reads the own-32 counts, each model's samples kept up to as many finished ones
as the reference has on the problem (\emph{matched}). A model \emph{adds} when it solves more problems beyond reach than
both decoding-only references and than the own-answer runs pooled, each 95\% paired interval above zero. It is
\emph{below} when its interval against a decoding-only reference lies below zero. Otherwise its addition is
\emph{bound}ed by the smaller upper end of its two intervals against the references. The distilled runs, read the same
way, show what the test can detect on each set.

\paragraph{Single-model flag and single-sample reliability.} A model is \emph{flagged} when its coverage is lower than every
fix's own, each fix scored against the other seven (95\% paired interval). On whole test sets, where the beyond-reach test
is blind, we compare one \emph{strict} sample (a cut answer counts as wrong) with one sample of the best decoding fix,
accuracy on the samples two settings both finish, and voting over thinking-off samples at no more tokens. Comparisons are
paired by problem, with 95\% bootstrap intervals over problems (10{,}000 resamples; 1{,}000 for flags).

\section{Results}
\label{sec:results}

\input{tables/rules}
\input{tables/ctrl}

\subsection{RQ1: Thinking off recovers most of the gain with no weight change}
\label{sec:fixes}

The edited model reaches \wEditMath\% on MATH-500 and \wEditGsm\% on GSM8K, against \wBaseMath\% and \wBaseGsm\% for the
unedited model (Table~\ref{tab:main}); the authors report \wAuthorsBaseMath\% and \wAuthorsEditMath\% on MATH-500, and
the edited model's card \citep{weightspacecard2026} \wCommunityEditMath\%. Switching thinking off while keeping the full
budget lifts accuracy to \wOffMath\% on MATH-500 and \wOffGsm\% on GSM8K, \wOffShareMath\% and \wOffShareGsm\% of the
gain we reproduce; on MATH-500 it exceeds the \wAuthorsEditMath\% the authors report for the edit. Without near-duplicates of
training problems the edit leads thinking off by \wCleanLead{} points (Appendix~\ref{app:protocol}).

\paragraph{Thinking-on fixes add answers at the model's own accuracy.} On MATH-500, $T = 1.0$, budget forcing and loop
stop lie within \wLineOnMax{} points of the line whose slope is the unedited model's accuracy on its own finished
answers, \wOwnFinAccMath\%; the textbook edit, the repetition penalty and the end-of-thinking bias sit \wLineTextbook,
\wLineRep{} and \wLineEot{} points below it, the agent's edit \wLineEdit{} and thinking off \wLineOff{}
(Figure~\ref{fig:completion}, left; textbook edits on practice problems, middle).

\paragraph{Released answers are no more accurate than the model's own.} An answer is \emph{released} when the unedited
model's sample is cut and the setting's sample of the same problem and index finishes. On the \wPairedMath{} MATH-500
problems where the unedited model sometimes finishes and sometimes loops, the edit's released answers are right
\wRelMath\% of the time and the model's own \wOwnMath\% (\wRelMinusOwnMath\pp{} \wRelMinusOwnMathCI); on GSM8K they are
less accurate, \wRelMinusOwnGsm\pp{} \wRelMinusOwnGsmCI. No fix releases more accurate answers than the model's own on
any test set (largest difference \wFixRelMax\pp, no interval above zero; Figure~\ref{fig:completion}, right).

\paragraph{On MMLU-Pro math, decoding recovers less of the gain.} Here the model's own answers, right \wOwnMmlu\% of the time, leave
room above them. The edit lifts accuracy from \wMmluBase\% to \wMmluEdit\%, above every decoding fix
(Table~\ref{tab:mmlu}), and thinking off recovers \wOffShareMmlu\% of the gain (\wRecStrictMmlu\% under strict scoring). Yet where the unedited model sometimes
finishes, the edit's released answers are right \wRelMmlu\% of the time (\wRelMinusOwnMmlu\pp{} \wRelMinusOwnMmluCI).

\subsection{RQ2: Nearly all of the edit's correct answers lie within the decoding reach, as do the controls'}

On the \wAlwaysMath{} MATH-500 and \wAlwaysGsm{} GSM8K problems where the unedited model always loops, the decoding reach
contains \wCovMath\% \wCovMathCI{} of the edit's \wEditCorrectMath{} correct answers and \wCovGsm\% \wCovGsmCI{} of its
\wEditCorrectGsm; the thinking-on reach alone contains \wCovOnMath\% and \wCovOnGsm\%. On MMLU-Pro math, on the
\wMmluNever{} questions the unedited model never finishes, the \wMmluRulerFixes{} fixes reach \wMmluCov\% of the edit's
\wMmluEditCorrect{} correct answers, where uniform guesses would reach \wMmluGuess\%; chance-corrected, \wMmluCorr\%
\wMmluCorrCI. This share locates the edit's answers but does not separate the controls: the own-answer runs keep
\wCovRelRangeMath\% of their correct MATH-500 answers within reach and the distilled runs \wCovDistRangeMath\%
(Table~\ref{tab:headline}), a contrast of \wRRulerDistBothMath{} \wRRulerDistBothMathCI{} points at $k = \wRulerSamples$
(Table~\ref{tab:rules}). Qwen3.5-2B, a model of more ability, leaves more beyond the 0.8B's reach (\wCovTwoBMath\% within
on MATH-500, \wTwoBReliableMath\% when matched to the edit's solved count; Table~\ref{tab:summary}), and every model's
beyond-reach share shrinks as the reach grows from 4 to 32 samples (Figure~\ref{fig:reach}). 

\begin{figure*}[t]
\centering
\includegraphics[width=\textwidth]{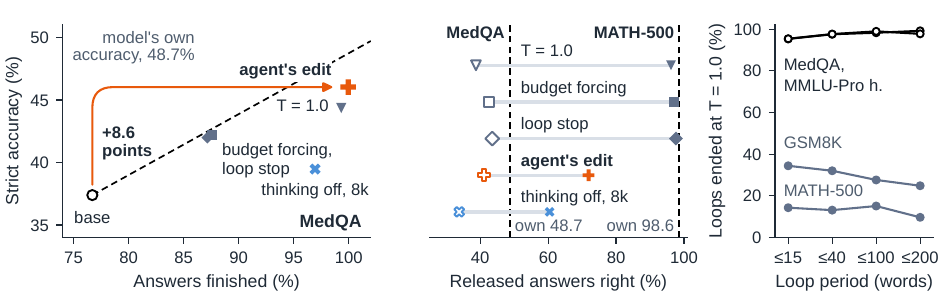}
\caption{Left, MedQA as in Figure~\ref{fig:completion}, left; middle, accuracy of every answer a fix releases
(dashed: the model's own, all finished answers); right, loops ended at $T{=}1.0$, by period. Filled: mathematics;
open: medicine.}
\label{fig:medicine}
\end{figure*}

\subsection{RQ3: Matched to decoding alone, the edit solves fewer problems beyond reach than thinking off on MATH-500 and GSM8K}

\paragraph{Against decoding alone.} Given 32 samples of its own, the edit solves \wPThirtyTwoEditBeyond{} of the
\wPThirtyTwoBeyondRuler{} MATH-500 problems beyond reach and \wOGsmSEdit{} of GSM8K's \wOGsmBeyond{}
(Table~\ref{tab:ctrl}). The unedited model with thinking off, given 32 further samples, solves \wBzMathSolved{} and
\wBzGsmSolved. Matched to its finished samples, the edit solves \wRaMathEditTo{} \wRaMathEditToCI{} and \wRaGsmEditTo{}
\wRaGsmEditToCI{} problems against it; against the unedited model sampled again, \wRaMathEditSa{} \wRaMathEditSaCI{} and
\wRaGsmEditSa{} \wRaGsmEditSaCI{} (Table~\ref{tab:refs}). On both sets the edit reads \emph{below}, also when cut samples
count (\wRaLooseMathEditTo{} \wRaLooseMathEditToCI{} on MATH-500).

\paragraph{The positive control marks what the test can see.} Read the same way, the distilled runs \emph{add} on GSM8K:
they solve more beyond reach than thinking off (\wRaGsmDistTo{} \wRaGsmDistToCI), than the model sampled again
(\wRaGsmDistSa{} \wRaGsmDistSaCI) and than the own-answer runs (\wRaGsmDistOwn{} \wRaGsmDistOwnCI). On MATH-500 they do
not exceed thinking off (\wRaMathDistTo{} \wRaMathDistToCI), so there the test bounds rather than detects: an addition the
size of distillation's would not be seen. The edit lies well below that reference on the same set.

\paragraph{Against the own-answer runs.} Every distilled run solves more beyond reach than every own-answer run on
MATH-500 and GSM8K (\wPEqPairsAbove{} of \wPEqPairs{} pairs each; Table~\ref{tab:rules}). On MATH-500 the edit's net excess
over the own-answer runs pooled is \wPEqEditRel{} \wPEqEditRelCI{} problems per run, so a net excess of up to
\wPEqEditRelHi{} is not excluded; against each own-answer run, \wPcEditOnlyLo--\wPcEditOnlyHi{} problems are solved by the
edit alone, \wPcRunOnlyLo--\wPcRunOnlyHi{} by the run alone and \wPcBothLo--\wPcBothHi{} by both. On GSM8K the edit solves
fewer than the own-answer runs (\wOGsmSEditRel{} \wOGsmSEditRelCI). The own-answer runs solve more GSM8K problems beyond reach than the model sampled again (matched,
\wZGsmRelM{} \wZGsmRelMCI) but not more than the model with thinking off (\wBzGsmRelM{} \wBzGsmRelMCI); on MMLU-Pro math
they exceed both references (\wBzMmluRelM{} \wBzMmluRelMCI{} against thinking off).

\paragraph{MMLU-Pro math: every count bounds the edit.} Under the chance-corrected count the edit is \wRaMmluEditSa{}
\wRaMmluEditSaCI{} against the model sampled again and \wRaMmluEditTo{} \wRaMmluEditToCI{} against thinking off, so an
addition of up to \wRaBoundMmluEdit{} of \wRaNMmlu{} questions is not excluded; with 80\% power this count detects an
excess of about \wRaDetMmluEditTo--\wRaDetMmluEditSa{} questions against these references. The binomial test, counting
cut samples and the fixed-$k$ count also bound the edit (at \wRaBoundBinMmluEdit, \wRaBoundLooseMmluEdit{} and
\wRaBoundFixkMmluEdit), although against the own-answer runs alone the first two place it above them
(\wRaBinMmluEditOwn{} \wRaBinMmluEditOwnCI{} and \wRaLooseMmluEditOwn{} \wRaLooseMmluEditOwnCI). Under this primary count the test bounds the distilled runs too; only the binomial test, a secondary count, detects them.

\paragraph{The accuracy criterion credits finishing.} By an accuracy criterion a model adds if, with thinking off, it
is more accurate than the unedited model with thinking off on the always-looping problems. Self-training sees only
answers within reach yet passes it on GSM8K and MMLU-Pro math (\wBtwoOffGsm{} \wBtwoOffGsmCI{} and \wBtwoOffMmlu{}
\wBtwoOffMmluCI{} points), and so does the edit on MATH-500 and MMLU-Pro math (\wCTwoMath{} \wCTwoMathCI{} and
\wCTwoMmlu{} \wCTwoMmluCI). On the samples both finish, however, the edit is no more accurate than the unedited model
(\wBfMathEditBase{} \wBfMathEditBaseCI{} and \wBfMmluEditBase{} \wBfMmluEditBaseCI; Table~\ref{tab:acc12}): its gains
come from finishing more answers.

\paragraph{The flag catches spiked-in additions of \wPowerGsm--\wPowerMath\%.} Correct answers beyond reach added to the
edit's at random are caught in 80\% of placements from an added share of \wPowerMath\%, \wPowerGsm\% and \wPowerMmlu\% on
MATH-500, GSM8K and MMLU-Pro math, and planted gold solutions lower coverage with their share (Figure~\ref{fig:reach},
right). On trained models it is a weak check: it flags \wFlagDistMath{} of the three distilled runs on MATH-500,
\wFlagDistGsm{} on GSM8K and \wFlagDistMmlu{} on MMLU-Pro math, and never the edit (Table~\ref{tab:rules}).

\subsection{RQ4: On whole test sets, one sample of the edit is more reliable than one sample of any decoding fix}

One strict sample of the edit beats one of the best decoding fix by \wResidMath\pp{} \wResidMathCI, \wResidGsm\pp{}
\wResidGsmCI{} and \wResidMmlu\pp{} \wResidMmluCI{} on MATH-500, GSM8K and MMLU-Pro math; on MATH-500 the lead is smaller
than the gap between the authors' run of the edit and ours (\wReconEdit\pp{} \wReconEditCI; Appendix~\ref{app:protocol}).
On the samples both finish the edit is right \wBothEditMath\%, \wBothEditGsm\% and \wBothEditMmlu\% of the time, against
\wBothOffMath\%, \wBothOffGsm\% and \wBothOffMmlu\% for thinking off. Where both finish, however, it is no more accurate
than budget forcing, loop stop or $T = 1.0$ (Table~\ref{tab:both}): its lead over them comes from finishing more answers
(Table~\ref{tab:main}). Ranked by agreement among its four samples, the edit's finished answers err less than thinking
off's at matched completion (\wRiskOffMath\pp{} \wRiskOffMathCI{} and \wRiskOffGsm\pp{} \wRiskOffGsmCI;
Appendix~\ref{app:reach}).

The lead is per sample, not per token: the edit spends \wTokRatioMath{} and \wTokRatioGsm{} times thinking off's tokens on
MATH-500 and GSM8K. At no more tokens than the edit uses, voting over \wMajEightMathJ{} thinking-off samples at 8{,}192
tokens beats it on MATH-500 (\wMajEightMathAcc\% against \wEditMath\%; \wMajEightMath\pp{} \wMajEightMathCI) and ties it
on GSM8K (\wMajEightGsm\pp{} \wMajEightGsmCI), and voting at the full budget ties it on both (\wMajMath\pp{} and
\wMajGsm\pp). On MMLU-Pro math the edit leads both: \wMajMmlu\pp{} \wMajMmluCI{} against the full budget and
\wMajEightMmlu\pp{} \wMajEightMmluCI{} against 8{,}192 tokens (Table~\ref{tab:majk}).

\subsection{RQ5: The pattern holds in medicine, on a second model and for frontier agents}

\input{tables/second}

\paragraph{Medicine.} The unedited model is right on \wOwnFinAccMed\% of the MedQA answers it finishes, so releasing
answers should bring little. Transferred unchanged, the edit gains \wMedEditGain\pp{} and sampling at $T = 1.0$
\wMedToneGain\pp; the difference is \wMedEditMinusTone\pp{} \wMedEditMinusToneCI{} (Table~\ref{tab:med};
Figure~\ref{fig:medicine}, left), and on MMLU-Pro health \wHealthEditMinusTone\pp{} \wHealthEditMinusToneCI. The edit's
released answers are right \wRelMed\% of the time, the model's own \wOwnMed\% (\wRelMinusOwnMed\pp{}
\wRelMinusOwnMedCI).

\paragraph{A second model.} Qwen3.5-2B loops less (always on \wSecAlwaysMath{} MATH-500 and \wSecAlwaysGsm{} GSM8K
problems). With its own eight fixes as the reach and its textbook edit as the target, all \wSecEditAnsGsm{} of the
edit's correct GSM8K answers lie within reach and it is \wSecEditFlaggedGsm, while Qwen3.5-4B
(\wSecFourBeyondGsm\% beyond reach) is \wSecFourFlaggedGsm, as is a spike-in from \wSecSpikeGsm\% (Table~\ref{tab:second}).

\paragraph{Frontier agents.} Given the authors' task without the test problems, answers or scorer, gpt-6.1-sol and
Claude Opus 5.5 found different edits (Appendix~\ref{app:harness}). On MATH-500 they gain \wAgentGptGain\pp{}
\wAgentGptGainCI{} and \wAgentClaudeGain\pp{} \wAgentClaudeGainCI{} over the unedited model (\wFrGptOffMath{} and
\wFrClaudeOffMath{} points from thinking off), and the decoding reach contains \wAgentGptCovMath\% and \wAgentClaudeCovMath\% of their correct answers where the
model always loops.

\section{Conclusion}
\label{sec:conclusion}
\label{sec:wsp}

\citet{takehi2026sparks} set their agent to improve the model's ``math reasoning capabilities'' and note that fixing
loops ``lets the model finish answers it could already work out'', but test this only on a model that rarely loops. On
their headline edit the note is the finding: thinking off recovers most of the gain, and, matched on finished samples, the unedited model with thinking off solves
more problems beyond the decoding reach than the edit on MATH-500 and GSM8K. What the edit keeps is
single-sample reliability on whole test sets, which voting over thinking-off samples matches or exceeds at no more tokens on
MATH-500 and GSM8K but not on MMLU-Pro math.  Their experiments on smaller networks and the
instruction-following gain are untested here. Before crediting an intervention on a looping model with new reasoning
ability, compare its gain with decoding fixes on the unedited model and count its correct answers beyond their decoding
reach against references that only decode, matched on finished samples.

\section*{Limitations}

Our main model is a single 0.8B model (on Qwen3.5-2B only the flag ran); the edits come from one published run and one
run per frontier agent. The test counts added ability only as answers beyond the reach of \wRulerSamples{} samples, the
shares still fall at \wPoolKMath{} samples (Table~\ref{tab:k}), and on MATH-500 it cannot detect an addition of
distillation's size. The own-answer controls rest on one answer source and solve more beyond reach than the model sampled again on GSM8K
and than both decoding-only references on MMLU-Pro math. The flag rule, the share contrast, the reading against the decoding-only
references and three registered rules (Table~\ref{tab:reg}) were fixed after part of what they read was known. Our
intervals resample problems, not serving setups: the authors', the edited model's card and our run give the edit
\wAuthorsEditMath--\wEditMath\% on MATH-500, a spread larger than its \wResidMath-point lead over the best fix.

\section*{Ethics Statement}

We use only public models and public benchmarks; no patient data is involved. The medical results describe model
behavior on exam-style questions and do not support clinical use. Licenses and compute are listed in
Appendix~\ref{app:compute}.

\section*{Acknowledgments}

We thank our colleagues for helpful discussions and feedback.

\paragraph{Writing assistance.} OpenAI Codex and Anthropic Claude assisted with language editing, consistency checks,
and figure-generation code.

\bibliography{custom}

\appendix
\renewcommand{\topfraction}{0.95}\renewcommand{\bottomfraction}{0.8}\renewcommand{\textfraction}{0.05}
\renewcommand{\floatpagefraction}{0.85}\renewcommand{\dbltopfraction}{0.95}\renewcommand{\dblfloatpagefraction}{0.85}
\setcounter{topnumber}{4}\setcounter{bottomnumber}{2}\setcounter{totalnumber}{6}\setcounter{dbltopnumber}{2}

A supplement released with the code (\coderepo) holds further analyses and per-run tables that this paper
does not rely on.

\section{Terms and Main Results in Detail}
\label{app:terms}

Table~\ref{tab:terms} collects the terms of \S\ref{sec:method}, and Table~\ref{tab:summary} gives
Table~\ref{tab:headline}'s quantities with MedQA and further references.

\begin{table}[t]
\centering
\footnotesize
\setlength{\tabcolsep}{3pt}
\begin{tabular}{@{}p{0.29\columnwidth}p{0.67\columnwidth}@{}}
\toprule
Term & Meaning \\
\midrule
Strict accuracy & Every cut answer counts as wrong \\
Released, own & Cut under the unedited model and finished under a setting (same problem and sample); finished by the unedited model \\
Always-looping & All four unedited samples of a problem are cut \\
Decoding reach & The always-looping problems that one of the unedited model's \wRulerSamples{} samples under its eight fixes solves (the protocol's own four finish none) \\
Within, beyond reach & Some reach sample finishes the answer's problem correctly (a cut sample never counts), or none does \\
Own-answer fine-tune & The 0.8B fine-tuned on its own correct answers under the end-of-thinking bias (\emph{bias answers}, within reach; \wPlantRelease{} MATH answers, \wSeedsNMath{} runs; \wAlignedNGsm{} more, \emph{data-aligned}, add \wAlignedGsmAns{} GSM8K answers on the distilled 0.8B's problems; Appendix~\ref{app:sft}) \\
Planted fine-tune & An own-answer fine-tune with gold solutions to problems beyond reach planted among its answers, at a known share \\
Self-training & The 0.8B fine-tuned on its own correct \emph{thinking-off answers} (\wNoffNMath{} runs), one round of STaR without rationalization \\
Coverage & The share of a model's correct answers within reach (per problem: of the problems it solves); the rest is its beyond-reach share \\
Own-32 count & Problems beyond reach that some finished sample of a model's own 32 solves (cut samples never count) \\
Thinking-on reach & From the six fixes that keep thinking on (24 samples) \\
Chance-corrected & $(c - g)/(1 - g)$, $g$ the coverage of uniform guesses among the options for the same finished samples \\
Coverage at $k$ & The unbiased pass@$k$ estimator \citep{chen2021codex} \\
\bottomrule
\end{tabular}
\caption{Terms used throughout.}
\label{tab:terms}
\end{table}
\input{tables/summary}

\section{Protocol and Reproduction Details}
\label{app:protocol}

We use the base model at a pinned revision and rebuild the agent's edit bit for bit from the published difference. Our
reproduction (\wEditMath\%, \wEditCutPctMath\% cut) differs from the authors' \wAuthorsEditMath\%
(\wAuthorsEditCut\% cut) by \wReconEdit\pp{} \wReconEditCI{} on the same problems, and the edited model's card
reports \wCommunityEditMath\% (\wCommunityEditCut\% cut). The authors' \wAuthorsBaseMath\% for the base model lies within our interval on the same
problems (\wReconBase\pp{} \wReconBaseCI). The loop detector checks the text every 512 thinking tokens for loop stop; its second
rule stands in for a repetition rule of the authors' that is not published. At
81{,}920 tokens, the edited model's answers that finish without a readable answer double (from \wNoAnswerEditMath{} to \wLongNoAnswer{} over four
samples). With 81{,}920 tokens, the base model's first sample on MATH-500 is right \wLongBaseAcc\% of the time, against
\wLongBaseCutAcc\% for the same answers cut at 32{,}768 tokens (a gain of \wLongBase\pp{} \wLongBaseCI: the loops are genuine), and the edit's first sample \wLongEditAcc\%; the extra
budget finishes \wLongEditFin{} of the edited model's \wLongEditReach{} answers that reach 32{,}768 tokens, and
\wLongEditRight{} of them are right (\wLongEditGain\pp{} \wLongEditGainCI).

Without the \wNearDup{} MATH-500 problems that are near-duplicates of MATH training problems (a shared 13-word window
or character $n$-gram cosine of at least 0.8), the edit reaches \wCleanEditAcc\% and leads thinking off at the full
budget by \wCleanLead\pp{} \wCleanLeadCI, against \wEditMinusOffMath{} on the full set. Given 81{,}920 tokens, the base
model's first sample finishes \wLongFinN{} of its \wLongCutN{} cut answers, and the edit's first sample still leads by
\wLongEditLead\pp{} \wLongEditLeadCI.

\paragraph{Forced answers.} Budget forcing and loop stop close the base model's thinking and let it answer within
\wForcedCap{} more tokens, in a new request, so vLLM's presence penalty does not count the thinking. Of the forced
answers on the three mathematics sets, \wForcedCutLo--\wForcedCutHi\% reach that cap, and \wForcedRepLo--\wForcedRepHi\%
of those still repeat there by the loop detector. The cap rarely binds an answer that would finish: forced answers that
finish take a median of \wForcedMedLo--\wForcedMedHi{} tokens, and \wForcedOverHalf\% of them use more than half of it. With a 16{,}384-token answer cap instead, budget forcing gets \wCapSixteenRight{} of \wCapTotal{} MATH-500 answers
right (\wCapSixteenFin\% finished), against \wCapFourRight{} (\wCapFourFin\%) at \wForcedCap.

\paragraph{Selection.} Hyperparameters were chosen on practice problems before any test result; the \emph{medical-direction edit}, the textbook edit's form with
$u$ from MedQA training questions, has its strength chosen on held-out MedQA training questions. Interval ends that
round to zero carry a further decimal, and an end that is exactly zero is printed 0; an interval excludes zero only if
its unrounded end does.

\section{Released Answers, Coverage and Matched Completion}
\label{app:reach}

Table~\ref{tab:med} gives the medical results, Table~\ref{tab:pair} the values behind Figure~\ref{fig:completion},
right, Table~\ref{tab:majk} majority voting, Table~\ref{tab:mix} the mixture control for both
donors (on GSM8K the distilled 0.8B's answers, \wMixDistSevenGsm\% beyond the seven fixes other than thinking off, add
\wMixDistEditAddedGsm\% at \wMixTwoBEditMaxGsm\% and are flagged in \wMixDistEditFiftyGsm\% of draws), Table~\ref{tab:both} the edit against every fix on the
samples both finish (on the problems that the edit and thinking off finish on every sample, \wBothFourEditMath\% against
\wBothFourOffMath\% on MATH-500 and \wBothFourEditGsm\% against \wBothFourOffGsm\% on GSM8K), and Table~\ref{tab:mmlu} every fix on MMLU-Pro math, where, on the questions the base model never
finishes, the edit's strict lead over the best of all \wMmluNeverFixes{} fixes is \wResidMmluNever\pp{}
\wResidMmluNeverCI.

\paragraph{The controls.} On always-looping problems thinking off at the full budget answers
\wOffOverDistLoMath--\wOffOverDistHiMath\%, \wOffOverDistLoGsm--\wOffOverDistHiGsm\% and
\wOffOverDistLoMmlu--\wOffOverDistHiMmlu\% as many correctly as the distilled runs (MATH-500, GSM8K, MMLU-Pro math).
Spike-ins add at most four answers per problem, at random, in 40 placements per share. Thinking off at the full budget
binds the flag: a flagged model must leave more beyond the other seven fixes than
thinking off's own answers do. Against those seven, thinking off's own answers on MATH-500 are \wSevenOffOwnMath\% covered and Qwen3.5-2B's \wSevenTwoBMath\% (GSM8K: \wSevenOffOwnGsm\% and
\wSevenTwoBGsm\%).

\paragraph{Self-training with thinking off.} Run with thinking off at the full budget, the two self-training runs are
right on \wBtwoOffAccMath\%, \wBtwoOffAccGsm\% and \wBtwoOffAccMmlu\% of the samples on MATH-500's and GSM8K's
always-looping problems and MMLU-Pro math's never-finished questions, the base model with thinking off on
\wBtwoBaseMath\%, \wBtwoBaseGsm\% and \wBtwoBaseMmlu\%; with thinking on, the authors'
protocol, on \wBtwoOnAccMath\%, \wBtwoOnAccGsm\% and \wBtwoOnAccMmlu\%. The reading calls the recipe self-training when the averaged paired difference lies above zero on at least two of the three sets.
On MMLU-Pro math the self-training runs leave \wMmluNoffShareLo--\wMmluNoffShareHi\% beyond reach (chance-corrected),
and one of them is flagged (\wMmluNoffFlag{} \wMmluNoffFlagCI{} points), a possible detection of what it adds.

\paragraph{Each model's own 32 samples.} On MATH-500's \wAlwaysMath{} always-looping problems we brought the edit, the
three distilled runs and the six own-answer runs from 4 to 32 samples under the authors' protocol and counted the
problems that some finished sample of a model solves and no reach sample does (\wPThirtyTwoBeyondRuler{} problems lie beyond the decoding reach; Table~\ref{tab:pass32}); as for the reach, a cut sample never counts. With 32 samples the
distilled runs solve \wPEqDistBeyondLo--\wPEqDistBeyondHi{} of them and the own-answer runs
\wPEqRelBeyondLo--\wPEqRelBeyondHi. Paired over problems (10{,}000 replicates over the \wAlwaysMath{} problems, the same
draws for every contrast), \wPEqPairsAbove{} of the \wPEqPairs{} distilled-minus-own-answer pairs lie above zero
(\wPEqFourPairsAbove{} with each model's first four samples); the distilled runs pooled minus the own-answer runs
pooled is \wPEqDistRel{} \wPEqDistRelCI{} problems per run (\wPEqFourDistRel{} \wPEqFourDistRelCI{} with four), the
edit minus the own-answer runs pooled \wPEqEditRel{} \wPEqEditRelCI{} and minus the distilled runs pooled
\wPEqEditDist{} \wPEqEditDistCI. The edit lies nominally above the own-answer runs under all four reaches and above all
\wPEqEditRelNomAbove{} single runs; its interval includes zero for \wPEqEditRelZero{} of 6
own-answer runs and lies below zero for \wPEqEditDistBelow{} of 3 distilled runs. Counting every right sample, cut or
not, only the edit changes, to \wPEqLooseEdit: its difference from the own-answer runs pooled becomes
\wPEqLooseEditRel{} \wPEqLooseEditRelCI, from the distilled runs \wPEqLooseEditDist{} \wPEqLooseEditDistCI. The edit-minus-own-answer contrast has a standard error of \wPEqSE{} problems: \wPEqPowerHalf\% power against an addition
of half the distilled advantage (\wPEqHalfDist), 80\% power from \wPEqDetectEighty. As a share of each model's correct
answers, which does not depend on how many it finishes, the edit's \wPEqShareEdit\% \wPEqShareEditCI{} lies
\wPEqShareEditRel{} \wPEqShareEditRelCI{} points from the own-answer runs pooled
(\wPEqShareRelLo--\wPEqShareRelHi\%) and \wPEqShareEditDist{} \wPEqShareEditDistCI{} from the distilled runs pooled
(\wPEqShareDistLo--\wPEqShareDistHi\%; distilled minus own-answer \wPEqShareDistRel{} \wPEqShareDistRelCI). The
32-sample rule for MATH-500 was set when the first pair and the edit's differences from it were known (\wPThirtyTwoDistRel{}
\wPThirtyTwoDistRelCI, \wPThirtyTwoEditRel{} \wPThirtyTwoEditRelCI{} and \wPThirtyTwoEditDist{} \wPThirtyTwoEditDistCI).

\paragraph{Own 32 samples on GSM8K and MMLU-Pro math.} The same design runs on the problems beyond the decoding reach: \wOGsmBeyond{} of GSM8K's \wAlwaysGsm{} always-looping problems and \wOMmluBeyond{}
of MMLU-Pro math's \wMmluNever{} never-finished questions. On GSM8K strict and loose counting agree (the edit
\wOGsmSEdit, own-answer \wOGsmSRelLo--\wOGsmSRelHi, distilled \wOGsmSDistLo--\wOGsmSDistHi; the edit is below the own-answer runs under both counts). On MMLU-Pro math four counts are reported: beyond chance (primary), the binomial test, the fixed-$k$
count and counting cut samples. Beyond chance, every model is below zero and the
controls do not separate (\wOMmluSDistRel{} \wOMmluSDistRelCI): with up to 32 finished samples and ten options,
uniform guesses solve a question with probability up to $1 - 0.9^{32} \approx 0.97$, and the edit and the distilled
runs, which finish more samples, are discounted most. The binomial test
gives the edit \wOMmluBEdit, own-answer \wOMmluBRelLo--\wOMmluBRelHi{} and distilled \wOMmluBDistLo--\wOMmluBDistHi{}
(the controls separate in \wOMmluBPairsAbove{} of \wPEqPairs{} pairs; against the decoding-only references, matched, the edit is bound and the distilled runs add; Table~\ref{tab:rules}). Counting cut samples with a readable
option too, every pair separates (\wOMmluLPairsAbove{} of \wPEqPairs) and the edit lies above the own-answer runs (\wOMmluLEditRel{}
\wOMmluLEditRelCI; against distilled \wOMmluLEditDist{} \wOMmluLEditDistCI), though there the own-answer runs' many
cut samples raise their chance level most. The unedited model sampled again solves \wOMathRefSolved, \wOGsmRefSolved{} and \wOMmluRefSolvedRaw{} of these
problems on MATH-500, GSM8K and MMLU-Pro math (\wOMmluRefSolved{} beyond chance), most of them with thinking off; on
every set fewer than every own-answer run solves with 32 samples of its own (Table~\ref{tab:ctrl}). Under the
fixed-$k$ count, which requires per question the right answers a binomial test asks of 32 samples (7 of 32 for ten
options), \wOMmluKPairsAbove{} of \wPEqPairs{} pairs separate and the edit lies \wOMmluKEditRel{} \wOMmluKEditRelCI{}
above the own-answer runs. Matched to the decoding-only references, the edit is bound under every one of these
counts (Table~\ref{tab:rules}).

\paragraph{The two references.} Table~\ref{tab:refs} gives every count against both. The unedited model with thinking off
takes samples 4--35 (0--3 belong to the reach) and finishes \wBzMathFin, \wBzGsmFin{} and \wBzMmluFin{} of 32 samples per
problem on MATH-500, GSM8K and MMLU-Pro math.

\input{tables/pass32runs}
\input{tables/acc12}
\input{tables/own32runs}
\input{tables/reg}
\input{tables/refs}

\paragraph{Matched completion.} Every setting's finished answers are ranked by the selector, the share of a problem's
four samples that give the same answer \citep{wang2023selfconsistency}, ties broken at random (we report the
expectation). A risk--coverage curve plots the strict error of the answers kept against the share
kept \citep{geifman2017selective}. The edit and thinking off at the full budget both finish \wBothShareMath\%,
\wBothShareGsm\% and \wBothShareMmlu\% of all samples; budget forcing, loop stop and $T = 1.0$ finish fewer with the edit
(\wBothShareOnLoMath--\wBothShareOnHiMath\% on MATH-500, \wBothShareOnLoGsm--\wBothShareOnHiGsm\% on GSM8K). At matched
completion the edit errs less than thinking off by \wRiskOffMath\pp{} \wRiskOffMathCI{} and \wRiskOffGsm\pp{}
\wRiskOffGsmCI; against $T = 1.0$, budget forcing and loop stop the gaps are \wRiskLoopMath{} to \wRiskToneMath{} and
\wRiskToneGsm{} to \wRiskLoopGsm{} points, and on GSM8K $T = 1.0$, loop stop and the prompt err less averaged over
shares kept. A perfect selector over thinking off's answers, keeping as many as the edit finishes,
would be right on \wOracleOffMath\% and \wOracleOffGsm\% of those kept, against \wEditKeptMath\% and \wEditKeptGsm\% of
the edit's finished answers.

\paragraph{Majority voting.} Thinking off at the full budget, voting over two samples at \wMajMathTok{} tokens per
problem, ties the edit's single sample at \wMajMathEditTok{} on MATH-500 (edit \wMajMath{} \wMajMathCI) and beats it
over three at \wMajMathThreeTok{} (\wMajMathThree{} \wMajMathThreeCI); on GSM8K, over four samples at \wMajGsmTok{}
against \wMajGsmEditTok, it ties it (\wMajGsm{} \wMajGsmCI; Table~\ref{tab:majk}).

\paragraph{Medicine.} At $T = 1.0$, \wToneBreakMed\% of the MedQA samples that loop under the base model finish,
against \wToneBreakMath\% on MATH-500, at every loop period (Figure~\ref{fig:medicine}, right).

\input{tables/med}

\input{tables/pair}
\input{tables/majk}
\input{tables/kshare}
\input{tables/runs}
\input{tables/both}
\input{tables/mix}
\input{tables/mmlu}

\section{Other Models}
\label{app:scale}

Two sets of 200 level-5 MATH training problems, one sample each, serve the other models: the first for the loop screen
of larger models and for R1-Distill's edit direction, the second for selecting R1-Distill's edit; the loop screen also
uses AIME 2024--2025 (60 problems).

Qwen3.5-2B's textbook edit is picked on the 200 practice problems among layer 15 or layers 15 and 19 at strengths 3, 6
and 9 (best: layer 15, strength 6; \wTwoBPickGain\pp). DeepSeek-R1-Distill-Qwen-1.5B's textbook edit takes its direction from
greedy answers on the first set of level-5 practice problems (layers 14, 17 and 21 of 28); on the second set no edit
beats the base model (best \wRoneSelBest\% against \wRoneSelBase\%), and the best edit (layers 17 and 21, strength 9) is the one run on
the test set, where it makes \wRoneNewCut{} problems that the base model finishes run to the budget. On the 200 practice problems, Qwen3.5-4B and 9B cut \wLargeCutPracLo--\wLargeCutPracHi\% of answers. Under the authors' protocol and the model
card's sampling they hit the budget on \wLargeCutLfiveLo--\wLargeCutLfiveHi\% of level-5 MATH practice problems and \wLargeCutAimeLo--\wLargeCutAimeHi\% of AIME 2024--2025
problems; of these \wLargeCut{} cut answers, \wLargeLoop{} repeat. On the \wRoneN{} test problems where the
base model loops and the edit finishes, the edit is right \wRoneRel\% of the time and the recommended temperature
\wRoneCardRel\% (\wRoneRelGap{} \wRoneRelGapCI); the base model is right on \wRoneOwn\% of the problems it finishes
itself, a different set.

\paragraph{The distilled 0.8B.}\label{app:sft} Its training problems are \wSftPoolMath{} MATH and \wSftPoolGsm{} GSM8K
training problems drawn at random from the practice pool after removing near-duplicates of MATH-500 and GSM8K test problems
(the rule of Appendix~\ref{app:protocol}). Qwen3.5-9B answers each once under the protocol with a \wSftTeacherCap-token budget, and its
\wSftExamples{} finished, correct solutions are the training set. The 0.8B model is fully fine-tuned with \wDistNGsm{}
training seeds (the data order), on an H200, with the loss on the teacher's tokens only: AdamW at a learning rate of \wSftLr, \wSftWarmup{} warm-up steps then a cosine
schedule, \wSftAccum{} examples per step, \wSftSteps{} steps (\wSftEpochs{} epochs). It is evaluated under the protocol
like every other setting.

\paragraph{The planted fine-tunes.} The own-answer data are the base model's answers to the \wPlantPool{} MATH problems of
the same practice pool under the end-of-thinking bias, one sample each within 8{,}192 tokens; its \wPlantRelease{}
finished, correct answers are kept. Planted problems are nested random draws from the \wPlantCandidates{}
always-looping MATH-500 problems that none of the \wPlantAllSamples{} samples of the protocol, the reach and seven other base
decodings (other budgets, temperatures and prompts) solves and whose gold solution from the MATH-500 release passes the
grader; each gold solution enters eight times, after an empty thinking block, the form most own answers take.
Training follows the distilled 0.8B's recipe for three epochs, and each model is evaluated under the protocol on the
always-looping problems; coverage falls with the planted share (slope \wPlantSlope{} \wPlantSlopeCI). Three more own-answer runs differ only in the training seed (the
data order); two of them trained on a workstation GPU with the reference implementation of the linear-attention
layers, the others on an H200 with fast kernels. On MATH-500 their shares do not split by machine (workstation
\wRunsLocalMath\%, H200 \wRunsHtwoMath\%); on GSM8K the workstation runs sit above the H200 runs (\wRunsLocalGsm\%
against \wRunsHtwoGsm\%), so the spread there is run-to-run, kernel included (Table~\ref{tab:k}). The \wAlignedNGsm{} data-aligned
own-answer runs (seeds 1 and 2, on an H200) add the base model's answers to the distilled 0.8B's \wSftPoolGsm{} GSM8K
problems under the same bias and filter (\wAlignedGsmAns{} kept). They differ from the distilled 0.8B only in their
answers: the base model's own correct ones (\wAlignedTotal) instead of the 9B's (\wSftExamples). The fine-tunes on
\emph{thinking-off answers}, \wNoffNMath{} runs (seeds 1 and 2, on the workstation GPU), learn the base model's answers
to the pool's MATH problems with thinking off (one sample within 8{,}192 tokens), finished and correct, after an empty thinking block
(\wNoffKept{} kept); every answer is within reach, since thinking off is one of the reach's fixes. A model's dose counts its correct answers on its planted problems
over all its correct answers there; intervals resample problems, the same replicates for every model. The slope falls short of $-1$ because the other answers' beyond-reach share drops at the two
largest doses (\wPlantOthersMid\% and \wPlantOthersTop\%, from \wPlantUnplantedZero\%), partly as plants take some of
the problems where they lie beyond reach: \wPlantZeroOnTop{} of the control's \wPlantBeyondZero{}
such answers lie on the largest dose's plants, and on the problems left unplanted the own-answer runs have
\wSeedsUnplantedLo--\wSeedsUnplantedHi{} beyond-reach answers and that fine-tune \wPlantOthersAnsTop.

\section{Frontier Agents: Runs and Intervals}
\label{app:harness}

Claude Opus 5.5 ran as \texttt{claude-opus-5-5} through the Claude Code command line and gpt-6.1-sol as
\texttt{openai.gpt-6.1-sol} on Amazon Bedrock through the Codex command line, both at maximum reasoning effort with the
providers' default sampling, their own tools disabled, in November 2026. The agents see the weights, the tokenizer and practice problems with answers (MATH and GSM8K training problems, AIME
1983--2023); the test problems, their answers and the scorer are not available to them. The authors' agent setup executes at most
\wAgentFwdBudget{} forward passes (a failed request is not counted) and 300 Python programs of at most 900 seconds each;
after the last pass the agent submits its edit. Both agents used all \wAgentFwdBudget{} passes; gpt-6.1-sol ran
\wAgentGptRuns{} Python programs and changed \wFrGptTensors{} tensor, Claude Opus 5.5 ran \wAgentClaudeRuns{} and changed
\wFrClaudeTensors. gpt-6.1-sol took \wFrGptTurns{} turns (\wFrGptIn{} input and \wFrGptOut{} output tokens), Claude Opus 5.5
\wFrClaudeTurns{} (\wFrClaudeIn{} and \wFrClaudeOut); the authors' run took \wAuthorsIn{} and \wAuthorsOut.
Without the \wNearDup{} near-duplicate MATH-500 problems, gpt-6.1-sol and Claude Opus 5.5 gain \wFrGptCleanBase{} and
\wFrClaudeCleanBase{} points over the base model, against thinking off at the full budget \wFrGptCleanOff{} and
\wFrClaudeCleanOff.

\paragraph{Leakage audit.} Both agents generated only on practice problems (\wAuditGptPractice{} distinct problems for
gpt-6.1-sol, \wAuditClaudePractice{} for Claude Opus 5.5); \wAuditGptNearDup{} and \wAuditClaudeNearDup{} of them are
among the \wAuditPoolNearDup{} training problems that are near-duplicates of a MATH-500 or GSM8K test problem, the
overlap behind the near-duplicate analysis. Neither wrote prompts of its own; their \wAuditGptForced{} and
\wAuditClaudeForced{} teacher-forced texts, their Python programs and their replies share no 13-word window with a
MATH-500, GSM8K or MedQA test problem. Claude Opus 5.5 drew some practice sets in MATH-500's level proportions, which
it stated from memory.

\section{The Agents' Task}
\label{app:task}

The system prompt is ``Answer the user's request.''; the first user message is the authors' task
\citep{takehi2026sparks} with their three notes to their agents appended (in full in the supplement). Its parts on the
goal and on repetition, verbatim:
\begin{quote}\small
Your job is to make it better at mathematics by EDITING ITS WEIGHTS DIRECTLY: find out what goes wrong inside the
network when it fails at math problems and repair that with targeted changes. [\dots] Repairing a specific failure
mechanism is in scope: for example, if the model often fails because it repeats itself and never finishes, finding and
fixing what drives that repetition counts. [\dots] Global knobs (making the whole output distribution sharper or
flatter) are not what we are looking for, and neither is making the model skip its thinking or close it early
regardless of the problem; repairing a specific failure inside the thinking (such as repetition) is in scope.
\end{quote}

\section{Compute and Licenses}
\label{app:compute}

All models are public (Qwen3.5: Apache 2.0; DeepSeek-R1-Distill: MIT) and all benchmarks are public. Experiments ran
on NVIDIA H200 GPUs and one RTX PRO 6000 workstation GPU.

\end{document}

%% file: numbers.tex
\newcommand{\wBaseMath}{7.6}
\newcommand{\wEditMath}{51.7}
\newcommand{\wOffMath}{48.6}
\newcommand{\wBaseCutMath}{93}
\newcommand{\wOffShareMath}{93}

\newcommand{\wEditMinusOffMath}{$+$3.1}

\newcommand{\wBaseGsm}{13.7}
\newcommand{\wEditGsm}{67.6}
\newcommand{\wOffGsm}{58.1}

\newcommand{\wOffShareGsm}{82}

\newcommand{\wMajMath}{$+$0.1}
\newcommand{\wMajMathCI}{[$-$2.6, 2.9]}
\newcommand{\wMajMathTok}{16.6k}
\newcommand{\wMajMathEditTok}{21.9k}

\newcommand{\wMajGsm}{$+$0.3}
\newcommand{\wMajGsmCI}{[$-$1.6, 2.2]}
\newcommand{\wMajGsmTok}{6.1k}
\newcommand{\wMajGsmEditTok}{11.9k}
\newcommand{\wMajEightMathJ}{four}
\newcommand{\wMajEightMath}{$-$4.4}
\newcommand{\wMajEightMathCI}{[$-$7.4, $-$1.5]}

\newcommand{\wMajEightMathAcc}{56.1}

\newcommand{\wMajEightGsm}{$+$0.9}
\newcommand{\wMajEightGsmCI}{[$-$0.9, 2.9]}

\newcommand{\wMajEightMmlu}{$+$5.5}
\newcommand{\wMajEightMmluCI}{[3.6, 7.4]}

\newcommand{\wCovMath}{97.7}
\newcommand{\wCovMathCI}{[96.3, 98.9]}
\newcommand{\wAlwaysMath}{426}
\newcommand{\wEditCorrectMath}{747}
\newcommand{\wCovTwoBMath}{92.0}

\newcommand{\wCovGsm}{99.3}
\newcommand{\wCovGsmCI}{[98.8, 99.6]}
\newcommand{\wAlwaysGsm}{934}
\newcommand{\wEditCorrectGsm}{2{,}138}

\newcommand{\wRelMath}{97.8}
\newcommand{\wOwnMath}{98.5}
\newcommand{\wRelMinusOwnMath}{$-$0.7}
\newcommand{\wRelMinusOwnMathCI}{[$-$5.1, 2.9]}
\newcommand{\wPairedMath}{68}

\newcommand{\wRelMinusOwnGsm}{$-$1.7}
\newcommand{\wRelMinusOwnGsmCI}{[$-$3.3, $-$0.2]}

\newcommand{\wRelMed}{42.2}
\newcommand{\wOwnMed}{42.9}
\newcommand{\wRelMinusOwnMed}{$-$0.7}
\newcommand{\wRelMinusOwnMedCI}{[$-$4.0, 2.6]}

\newcommand{\wMmluBase}{40.7}
\newcommand{\wMmluEdit}{62.4}
\newcommand{\wRelMmlu}{84.6}
\newcommand{\wOwnMmlu}{90.1}
\newcommand{\wRelMinusOwnMmlu}{$-$5.5}
\newcommand{\wRelMinusOwnMmluCI}{[$-$7.8, $-$3.3]}

\newcommand{\wRiskOffMath}{$+$7.5}
\newcommand{\wRiskOffMathCI}{[3.9, 10.9]}
\newcommand{\wRiskOffGsm}{$+$10.9}
\newcommand{\wRiskOffGsmCI}{[9.3, 12.7]}

\newcommand{\wLongBase}{$+$0.2}
\newcommand{\wLongBaseCI}{[0, 0.6]}
\newcommand{\wLongEditLead}{$+$46.6}
\newcommand{\wLongEditLeadCI}{[42.0, 51.0]}

\newcommand{\wMedEditMinusTone}{$+$1.5}
\newcommand{\wMedEditMinusToneCI}{[$-$0.04, 3.2]}

\newcommand{\wAgentGptGain}{$+$35.1}
\newcommand{\wAgentGptGainCI}{[31.8, 38.4]}

\newcommand{\wAgentClaudeGain}{$+$25.7}
\newcommand{\wAgentClaudeGainCI}{[22.8, 28.5]}

\newcommand{\wAgentClaudeRuns}{295}

\newcommand{\wAgentGptRuns}{66}

\newcommand{\wAgentFwdBudget}{50}
\newcommand{\wAuditClaudePractice}{5{,}246}
\newcommand{\wAuditClaudeNearDup}{114}
\newcommand{\wAuditClaudeForced}{236}
\newcommand{\wAuditGptPractice}{3{,}586}
\newcommand{\wAuditGptNearDup}{70}
\newcommand{\wAuditGptForced}{34}
\newcommand{\wAuditPoolNearDup}{343}

\newcommand{\wCleanLead}{$+$2.9}
\newcommand{\wCleanLeadCI}{[0.1, 5.6]}

\newcommand{\wCleanEditAcc}{48.2}
\newcommand{\wNearDup}{111}

\newcommand{\wTokRatioMath}{2.6}

\newcommand{\wTokRatioGsm}{7.8}

\newcommand{\wLongCutN}{454}
\newcommand{\wLongFinN}{1}
\newcommand{\wOwnFinAccMath}{98.6}
\newcommand{\wOwnFinAccMed}{48.7}

\newcommand{\wFixRelMax}{$+$2.9}

\newcommand{\wMmluNever}{842}

\newcommand{\wMmluEditCorrect}{1{,}465}

\newcommand{\wMmluCov}{96.9}

\newcommand{\wMmluGuess}{75.2}
\newcommand{\wMmluCorr}{87.5}
\newcommand{\wMmluCorrCI}{[83.2, 91.5]}

\newcommand{\wRiskToneMath}{$+$1.6}

\newcommand{\wRiskLoopMath}{$+$1.1}

\newcommand{\wRiskToneGsm}{$+$3.0}

\newcommand{\wRiskLoopGsm}{$+$3.3}

\newcommand{\wMedEditGain}{$+$3.5}

\newcommand{\wMedToneGain}{$+$2.0}

\newcommand{\wHealthEditMinusTone}{$+$0.8}
\newcommand{\wHealthEditMinusToneCI}{[$-$0.8, 2.5]}

\newcommand{\wToneBreakMed}{97.9}
\newcommand{\wToneBreakMath}{13.9}

\newcommand{\wFrGptTurns}{124}
\newcommand{\wFrGptIn}{125.7M}
\newcommand{\wFrGptOut}{3.9M}

\newcommand{\wFrGptTensors}{one}

\newcommand{\wFrGptOffMath}{$-$5.9 [$-$8.5, $-$3.4]}

\newcommand{\wFrGptCleanBase}{$+$32.8 [28.9, 36.5]}
\newcommand{\wFrGptCleanOff}{$-$6.0 [$-$9.0, $-$3.2]}
\newcommand{\wFrClaudeTurns}{355}
\newcommand{\wFrClaudeIn}{31.8M}
\newcommand{\wFrClaudeOut}{1.1M}

\newcommand{\wFrClaudeTensors}{four}

\newcommand{\wFrClaudeOffMath}{$-$15.4 [$-$18.0, $-$12.7]}

\newcommand{\wFrClaudeCleanBase}{$+$23.4 [20.3, 26.5]}
\newcommand{\wFrClaudeCleanOff}{$-$15.4 [$-$18.5, $-$12.5]}

\newcommand{\wLargeCut}{143}
\newcommand{\wLargeLoop}{4}

\newcommand{\wTwoBReliableMath}{93.8}

\newcommand{\wEditCutPctMath}{30.9}
\newcommand{\wLongBaseAcc}{9.6}
\newcommand{\wLongBaseCutAcc}{9.4}
\newcommand{\wLongEditAcc}{56.2}
\newcommand{\wLongEditFin}{69}
\newcommand{\wLongEditReach}{145}
\newcommand{\wLongEditRight}{25}
\newcommand{\wLongEditGain}{$+$5.0}
\newcommand{\wLongEditGainCI}{[3.2, 7.0]}
\newcommand{\wLongNoAnswer}{248}
\newcommand{\wNoAnswerEditMath}{123}

\newcommand{\wMajMathThree}{$-$4.7}
\newcommand{\wMajMathThreeCI}{[$-$7.5, $-$1.9]}
\newcommand{\wMajMathThreeTok}{24.9k}

\newcommand{\wRoneN}{55}
\newcommand{\wRoneRel}{81.8}
\newcommand{\wRoneCardRel}{72.7}
\newcommand{\wRoneOwn}{96.4}
\newcommand{\wRoneRelGap}{$+$9.1}
\newcommand{\wRoneRelGapCI}{[0, 20.0]}
\newcommand{\wSftExamples}{3{,}159}
\newcommand{\wSftEpochs}{three}

\newcommand{\wAuthorsBaseMath}{7.9}
\newcommand{\wAuthorsEditMath}{48.1}
\newcommand{\wAuthorsGainMath}{40.2}
\newcommand{\wAuthorsEditCut}{33.1}
\newcommand{\wReconBase}{$-$0.4}
\newcommand{\wReconBaseCI}{[$-$1.7, 1.0]}
\newcommand{\wReconEdit}{$+$3.6}
\newcommand{\wReconEditCI}{[1.7, 5.5]}
\newcommand{\wCommunityEditMath}{51.0}
\newcommand{\wCommunityEditCut}{31.6}
\newcommand{\wAuthorsEditShare}{0.016}
\newcommand{\wSftPoolMath}{3{,}000}
\newcommand{\wSftPoolGsm}{1{,}000}
\newcommand{\wSftTeacherCap}{12{,}288}
\newcommand{\wSftSteps}{294}
\newcommand{\wSftAccum}{32}
\newcommand{\wSftWarmup}{10}
\newcommand{\wSftLr}{$10^{-5}$}

\newcommand{\wOffShareMmlu}{56}

\newcommand{\wPowerMath}{8}

\newcommand{\wPowerGsm}{2}

\newcommand{\wPowerMmlu}{4}

\newcommand{\wMajMmlu}{$+$9.4}
\newcommand{\wMajMmluCI}{[7.9, 10.9]}

\newcommand{\wAgentGptCovMath}{99.7}

\newcommand{\wAgentClaudeCovMath}{99.3}

\newcommand{\wCovOnMath}{88.0}

\newcommand{\wCovOnGsm}{97.0}

\newcommand{\wLineOnMax}{0.4}
\newcommand{\wLineRep}{2.0}
\newcommand{\wLineTextbook}{1.5}
\newcommand{\wLineEot}{7.3}
\newcommand{\wLineEdit}{16.5}
\newcommand{\wLineOff}{36.2}

\newcommand{\wTwoBPickGain}{$+$3.5}
\newcommand{\wRoneSelBase}{53.0}
\newcommand{\wRoneSelBest}{51.0}
\newcommand{\wRoneNewCut}{38}
\newcommand{\wLargeCutPracLo}{2}
\newcommand{\wLargeCutPracHi}{2.5}
\newcommand{\wLargeCutLfiveLo}{4}
\newcommand{\wLargeCutLfiveHi}{10.5}
\newcommand{\wLargeCutAimeLo}{27}
\newcommand{\wLargeCutAimeHi}{40}

\newcommand{\wAuthorsIn}{86.8M}
\newcommand{\wAuthorsOut}{2.9M}

\newcommand{\wPlantPool}{3{,}000}
\newcommand{\wPlantRelease}{1{,}039}

\newcommand{\wPlantAllSamples}{61}
\newcommand{\wPlantCandidates}{99}

\newcommand{\wPlantBeyondZero}{13}
\newcommand{\wPlantSlope}{$-$0.90}
\newcommand{\wPlantSlopeCI}{[$-$0.94, $-$0.83]}

\newcommand{\wPlantUnplantedZero}{2.0}

\newcommand{\wMmluRulerFixes}{eight}

\newcommand{\wResidMath}{$+$3.1}
\newcommand{\wResidMathCI}{[0.6, 5.6]}

\newcommand{\wBothEditMath}{77.5}
\newcommand{\wBothOffMath}{65.2}

\newcommand{\wBothShareMath}{62.7}

\newcommand{\wBothFourEditMath}{90.7}
\newcommand{\wBothFourOffMath}{75.8}

\newcommand{\wResidGsm}{$+$9.3}
\newcommand{\wResidGsmCI}{[7.9, 10.8]}

\newcommand{\wBothEditGsm}{72.9}
\newcommand{\wBothOffGsm}{60.4}

\newcommand{\wBothShareGsm}{92.0}

\newcommand{\wBothFourEditGsm}{78.5}
\newcommand{\wBothFourOffGsm}{66.2}
\newcommand{\wRecStrictMmlu}{77}
\newcommand{\wResidMmlu}{$+$8.6}
\newcommand{\wResidMmluCI}{[7.1, 10.1]}

\newcommand{\wBothEditMmlu}{72.6}
\newcommand{\wBothOffMmlu}{60.5}

\newcommand{\wBothShareMmlu}{79.2}

\newcommand{\wResidLo}{3.1}
\newcommand{\wResidHi}{9.3}

\newcommand{\wResidMmluNever}{$+$7.9}
\newcommand{\wResidMmluNeverCI}{[5.8, 9.9]}
\newcommand{\wMmluNeverFixes}{8}
\newcommand{\wOracleOffMath}{70.2}
\newcommand{\wEditKeptMath}{74.7}
\newcommand{\wOracleOffGsm}{62.3}
\newcommand{\wEditKeptGsm}{72.6}
\newcommand{\wRulerSamples}{32}

\newcommand{\wPlantOthersMid}{0.6}

\newcommand{\wPlantOthersTop}{0.5}

\newcommand{\wPlantOthersAnsTop}{3}
\newcommand{\wPlantZeroOnTop}{7}

\newcommand{\wMixTwoBEditMaxGsm}{50}

\newcommand{\wForcedCap}{4{,}096}
\newcommand{\wForcedCutLo}{46}
\newcommand{\wForcedCutHi}{70}
\newcommand{\wForcedRepLo}{58}
\newcommand{\wForcedRepHi}{89}
\newcommand{\wForcedMedLo}{357}
\newcommand{\wForcedMedHi}{634}
\newcommand{\wForcedOverHalf}{1.1}

\newcommand{\wSecAlwaysMath}{97}

\newcommand{\wSecAlwaysGsm}{115}

\newcommand{\wSecEditAnsGsm}{167}

\newcommand{\wSecEditFlaggedGsm}{not flagged}
\newcommand{\wSecSpikeGsm}{7}

\newcommand{\wSecFourBeyondGsm}{5.8}

\newcommand{\wSecFourFlaggedGsm}{flagged}

\newcommand{\wCapSixteenFin}{33.7}
\newcommand{\wCapFourFin}{34.2}

\newcommand{\wCapSixteenRight}{661}
\newcommand{\wCapFourRight}{668}
\newcommand{\wCapTotal}{2{,}000}

\newcommand{\wPoolKMath}{57}
\newcommand{\wSeedsNMath}{four}

\newcommand{\wSeedsUnplantedLo}{5}
\newcommand{\wSeedsUnplantedHi}{10}
\newcommand{\wBothShareOnLoMath}{17.6}
\newcommand{\wBothShareOnHiMath}{32.3}

\newcommand{\wBothShareOnLoGsm}{34.5}
\newcommand{\wBothShareOnHiGsm}{57.3}

\newcommand{\wRunsHtwoMath}{2.0 and 2.4}
\newcommand{\wRunsLocalMath}{2.0 and 2.1}

\newcommand{\wRunsHtwoGsm}{1.1 and 1.1}
\newcommand{\wRunsLocalGsm}{1.3 and 1.7}

\newcommand{\wDistNGsm}{three}

\newcommand{\wAlignedNGsm}{two}

\newcommand{\wAlignedGsmAns}{564}
\newcommand{\wAlignedTotal}{1{,}603}

\newcommand{\wMmluNoffShareLo}{17.1}
\newcommand{\wMmluNoffShareHi}{17.2}

\newcommand{\wMmluNoffFlag}{$+$5.2}
\newcommand{\wMmluNoffFlagCI}{[0.7, 9.4]}

\newcommand{\wNoffNMath}{two}

\newcommand{\wNoffKept}{1{,}267}
\newcommand{\wOffOverDistLoMath}{86}
\newcommand{\wOffOverDistHiMath}{88}

\newcommand{\wOffOverDistLoGsm}{71}
\newcommand{\wOffOverDistHiGsm}{72}

\newcommand{\wOffOverDistLoMmlu}{89}
\newcommand{\wOffOverDistHiMmlu}{90}

\newcommand{\wFlagDistMath}{none}
\newcommand{\wRRulerDistBothMath}{$+$0.7}
\newcommand{\wRRulerDistBothMathCI}{[$-$0.4, 1.8]}

\newcommand{\wFlagDistGsm}{three}

\newcommand{\wFlagDistMmlu}{none}

\newcommand{\wMixDistSevenGsm}{4.5}
\newcommand{\wMixDistEditAddedGsm}{1.7}
\newcommand{\wMixDistEditFiftyGsm}{28}

\newcommand{\wPThirtyTwoBeyondRuler}{101}

\newcommand{\wPThirtyTwoEditBeyond}{36}

\newcommand{\wPThirtyTwoDistRel}{$+$21}
\newcommand{\wPThirtyTwoDistRelCI}{[11, 32]}

\newcommand{\wPThirtyTwoEditRel}{$+$6}
\newcommand{\wPThirtyTwoEditRelCI}{[$-$3, 15]}

\newcommand{\wPThirtyTwoEditDist}{$-$15}
\newcommand{\wPThirtyTwoEditDistCI}{[$-$25, $-$5]}

\newcommand{\wPEqDistRel}{$+$19}
\newcommand{\wPEqDistRelCI}{[12, 27]}
\newcommand{\wPEqEditRel}{$+$6}
\newcommand{\wPEqEditRelCI}{[$-$2, 15]}
\newcommand{\wPEqEditDist}{$-$13}
\newcommand{\wPEqEditDistCI}{[$-$22, $-$4]}

\newcommand{\wPEqFourDistRel}{$+$6}
\newcommand{\wPEqFourDistRelCI}{[0.2, 12]}
\newcommand{\wPEqFourPairsAbove}{3}
\newcommand{\wPEqPairsAbove}{18}
\newcommand{\wPEqPairs}{18}

\newcommand{\wPEqEditRelZero}{6}
\newcommand{\wPEqEditDistBelow}{2}

\newcommand{\wPEqDistBeyondLo}{44}
\newcommand{\wPEqDistBeyondHi}{52}
\newcommand{\wPEqRelBeyondLo}{28}
\newcommand{\wPEqRelBeyondHi}{31}

\newcommand{\wPEqEditRelHi}{15}

\newcommand{\wPEqHalfDist}{9.6}
\newcommand{\wPEqEditRelNomAbove}{6}

\newcommand{\wBtwoOffAccMath}{42.0}

\newcommand{\wBtwoOnAccMath}{43.1}
\newcommand{\wBtwoBaseMath}{41.1}
\newcommand{\wBtwoOffGsm}{$+$2.0}
\newcommand{\wBtwoOffGsmCI}{[0.3, 3.7]}
\newcommand{\wBtwoOffAccGsm}{50.0}

\newcommand{\wBtwoOnAccGsm}{50.7}
\newcommand{\wBtwoBaseGsm}{48.0}
\newcommand{\wBtwoOffMmlu}{$+$1.9}
\newcommand{\wBtwoOffMmluCI}{[0.2, 3.5]}
\newcommand{\wBtwoOffAccMmlu}{37.5}

\newcommand{\wBtwoOnAccMmlu}{38.7}
\newcommand{\wBtwoBaseMmlu}{35.6}

\newcommand{\wPEqLooseEdit}{38}
\newcommand{\wPEqLooseEditRel}{$+$8}
\newcommand{\wPEqLooseEditRelCI}{[0.2, 17]}
\newcommand{\wPEqLooseEditDist}{$-$11}
\newcommand{\wPEqLooseEditDistCI}{[$-$20, $-$2]}

\newcommand{\wOMathRefSolved}{24}

\newcommand{\wOGsmSEditRel}{$-$9}
\newcommand{\wOGsmSEditRelCI}{[$-$16, $-$2]}

\newcommand{\wOGsmSEdit}{41}
\newcommand{\wOGsmSRelLo}{44}
\newcommand{\wOGsmSRelHi}{57}
\newcommand{\wOGsmSDistLo}{72}
\newcommand{\wOGsmSDistHi}{73}

\newcommand{\wOGsmBeyond}{101}
\newcommand{\wOGsmRefSolved}{28}

\newcommand{\wOMmluSDistRel}{$-$7}
\newcommand{\wOMmluSDistRelCI}{[$-$17, 3]}

\newcommand{\wOMmluLEditRel}{$+$21}
\newcommand{\wOMmluLEditRelCI}{[9, 32]}
\newcommand{\wOMmluLEditDist}{$-$5}
\newcommand{\wOMmluLEditDistCI}{[$-$16, 6]}

\newcommand{\wOMmluLPairsAbove}{18}

\newcommand{\wOMmluBPairsAbove}{16}

\newcommand{\wOMmluBEdit}{25}
\newcommand{\wOMmluBRelLo}{10}
\newcommand{\wOMmluBRelHi}{17}
\newcommand{\wOMmluBDistLo}{27}
\newcommand{\wOMmluBDistHi}{34}

\newcommand{\wOMmluKEditRel}{$+$21}
\newcommand{\wOMmluKEditRelCI}{[13, 31]}

\newcommand{\wOMmluKPairsAbove}{17}

\newcommand{\wOMmluBeyond}{185}
\newcommand{\wOMmluRefSolved}{$-$18}
\newcommand{\wOMmluRefSolvedRaw}{66}

\newcommand{\wCTwoMath}{$+$3.7}
\newcommand{\wCTwoMathCI}{[1.0, 6.5]}

\newcommand{\wCTwoMmlu}{$+$3.7}
\newcommand{\wCTwoMmluCI}{[1.8, 5.6]}

\newcommand{\wBzMathSolved}{55}
\newcommand{\wBzMathFin}{21.2}

\newcommand{\wBfMathEditBase}{$-$0.6}
\newcommand{\wBfMathEditBaseCI}{[$-$3.6, 2.5]}

\newcommand{\wZGsmRelM}{$+$14}
\newcommand{\wZGsmRelMCI}{[7, 22]}

\newcommand{\wBzGsmSolved}{59}
\newcommand{\wBzGsmFin}{31.8}

\newcommand{\wBzGsmRelM}{$-$10}
\newcommand{\wBzGsmRelMCI}{[$-$19, 1]}

\newcommand{\wBzMmluFin}{23.4}

\newcommand{\wBzMmluRelM}{$+$16}
\newcommand{\wBzMmluRelMCI}{[4, 28]}

\newcommand{\wBfMmluEditBase}{$-$0.6}
\newcommand{\wBfMmluEditBaseCI}{[$-$2.6, 1.5]}

\newcommand{\wPEqSE}{4.2}
\newcommand{\wPEqPowerHalf}{63}
\newcommand{\wPEqDetectEighty}{12}
\newcommand{\wPEqShareEdit}{1.9}
\newcommand{\wPEqShareEditCI}{[1.1, 3.0]}
\newcommand{\wPEqShareRelLo}{1.7}
\newcommand{\wPEqShareRelHi}{1.9}
\newcommand{\wPEqShareDistLo}{2.6}
\newcommand{\wPEqShareDistHi}{2.9}
\newcommand{\wPEqShareDistRel}{$+$1.0}
\newcommand{\wPEqShareDistRelCI}{[0.3, 1.7]}
\newcommand{\wPEqShareEditRel}{$+$0.2}
\newcommand{\wPEqShareEditRelCI}{[$-$0.5, 0.9]}
\newcommand{\wPEqShareEditDist}{$-$0.8}
\newcommand{\wPEqShareEditDistCI}{[$-$1.5, $-$0.1]}

\newcommand{\wSevenOffOwnMath}{90.9}
\newcommand{\wSevenTwoBMath}{84.1}

\newcommand{\wSevenOffOwnGsm}{97.9}
\newcommand{\wSevenTwoBGsm}{94.7}
\newcommand{\wCovRelRangeMath}{97.5--98.7}
\newcommand{\wCovDistRangeMath}{97.2--97.4}

\newcommand{\wRaMathEditSa}{$+$1}
\newcommand{\wRaMathEditSaCI}{[$-$9, 11]}

\newcommand{\wRaMathEditTo}{$-$21}
\newcommand{\wRaMathEditToCI}{[$-$33, $-$9]}

\newcommand{\wRaMathDistTo}{$-$6}
\newcommand{\wRaMathDistToCI}{[$-$16, 4]}

\newcommand{\wRaLooseMathEditTo}{$-$19}
\newcommand{\wRaLooseMathEditToCI}{[$-$31, $-$8]}

\newcommand{\wRaGsmEditSa}{$-$2}
\newcommand{\wRaGsmEditSaCI}{[$-$10, 6]}

\newcommand{\wRaGsmEditTo}{$-$18}
\newcommand{\wRaGsmEditToCI}{[$-$30, $-$7]}

\newcommand{\wRaGsmDistSa}{$+$28}
\newcommand{\wRaGsmDistSaCI}{[19, 39]}

\newcommand{\wRaGsmDistTo}{$+$13}
\newcommand{\wRaGsmDistToCI}{[3, 23]}

\newcommand{\wRaGsmDistOwn}{$+$23}
\newcommand{\wRaGsmDistOwnCI}{[16, 31]}

\newcommand{\wRaBoundMmluEdit}{8}
\newcommand{\wRaMmluEditSa}{$+$5}
\newcommand{\wRaMmluEditSaCI}{[$-$11, 21]}
\newcommand{\wRaDetMmluEditSa}{22}
\newcommand{\wRaMmluEditTo}{$-$6}
\newcommand{\wRaMmluEditToCI}{[$-$20, 8]}
\newcommand{\wRaDetMmluEditTo}{20}

\newcommand{\wRaBoundLooseMmluEdit}{2}

\newcommand{\wRaLooseMmluEditOwn}{$+$21}
\newcommand{\wRaLooseMmluEditOwnCI}{[9, 32]}

\newcommand{\wRaBoundBinMmluEdit}{10}

\newcommand{\wRaBinMmluEditOwn}{$+$12}
\newcommand{\wRaBinMmluEditOwnCI}{[4, 21]}

\newcommand{\wRaBoundFixkMmluEdit}{13}

\newcommand{\wRaNMmlu}{185}
\newcommand{\wPcEditOnlyLo}{13}
\newcommand{\wPcEditOnlyHi}{19}

\newcommand{\wPcRunOnlyLo}{7}
\newcommand{\wPcRunOnlyHi}{11}

\newcommand{\wPcBothLo}{17}
\newcommand{\wPcBothHi}{23}

%% file: tables/headline.tex
\begin{table*}[t]
\centering
\footnotesize
\setlength{\tabcolsep}{4pt}
\begin{tabular}{@{}llll@{}}
\toprule
& MATH-500 & GSM8K & MMLU-Pro math \\
\midrule
\multicolumn{4}{@{}l}{\emph{RQ1: the gain, and how much of it thinking off recovers with no weight change}} \\
Accuracy, base $\to$ agent's edit (\%) & 7.6 $\to$ 51.7 & 13.7 $\to$ 67.6 & 40.7 $\to$ 62.4 \\
Gain recovered by thinking off, authors'\,/\,strict (\%) & 93\,/\,93 & 82\,/\,83 & 56\,/\,77 \\
\multicolumn{4}{@{}l}{\emph{RQ2: correct answers on always-looping problems within the decoding reach (\%)}} \\
Agent's edit & 97.7 [96.3, 98.9] & 99.3 [98.8, 99.6] & 87.5 [83.2, 91.5]$^{c}$ \\
Own-answer runs\,/\,distilled runs & 97.5--98.7\,/\,97.2--97.4 & 98.3--98.9\,/\,97.1--97.3 & 88.4--94.4\,/\,83.3--87.5$^{c}$ \\
\multicolumn{4}{@{}l}{\emph{RQ3: problems beyond reach solved with 32 samples, minus a decoding-only reference (matched)}} \\
Edit $-$ unedited model, sampled again & $+$1 [$-$9, 11] & $-$2 [$-$10, 6] & $+$5 [$-$11, 21]$^{c}$ \\
Edit $-$ unedited model, thinking off & $-$21 [$-$33, $-$9] & $-$18 [$-$30, $-$7] & $-$6 [$-$20, 8]$^{c}$ \\
Distilled runs $-$ unedited model, thinking off & $-$6 [$-$16, 4] & $+$13 [3, 23] & $+$9 [$-$5, 23]$^{c}$ \\
Reading, edit\,/\,distilled & below\,/\,$\le$4 & below\,/\,adds & $\le$8\,/\,$\le$23 \\
\multicolumn{4}{@{}l}{\emph{RQ4: whole test sets (points)}} \\
Edit $-$ best fix, one strict sample each & $+$3.1 [0.6, 5.6] & $+$9.3 [7.9, 10.8] & $+$8.6 [7.1, 10.1] \\
Edit $-$ thinking-off vote, full budget$^{v}$ & $+$0.1 [$-$2.6, 2.9] & $+$0.3 [$-$1.6, 2.2] & $+$9.4 [7.9, 10.9] \\
Edit $-$ thinking-off vote, 8k$^{v}$ & $-$4.4 [$-$7.4, $-$1.5] & $+$0.9 [$-$0.9, 2.9] & $+$5.5 [3.6, 7.4] \\
\bottomrule
\end{tabular}
\caption{Main results (Qwen3.5-0.8B; 4 samples per problem unless stated; 95\% intervals over problems). RQ3: each model's samples kept up to as many finished ones as the reference has on the problem; reading (\S\ref{sec:method}): \emph{adds}, \emph{below}, or bound, $\le$ the largest addition not excluded (problems). $^{v}$Majority vote over as many thinking-off samples as fit in the edit's tokens. $^{c}$Chance-corrected. Every quantity per set: Table~\ref{tab:summary}.}
\label{tab:headline}
\end{table*}

%% file: tables/main.tex
\begin{table}[t]
\centering
\small
\setlength{\tabcolsep}{2.6pt}
\begin{tabular}{@{}lrrrrrr@{}}
\toprule
& \multicolumn{3}{c}{MATH-500} & \multicolumn{3}{c}{GSM8K} \\
\cmidrule(lr){2-4}\cmidrule(l){5-7}
& Acc & Fin & Tok & Acc & Fin & Tok \\
\midrule
Base & 7.6 & 7.1 & 30.9 & 13.7 & 12.3 & 29.3 \\
Prompt only & 7.9 & 7.1 & 30.8 & 13.6 & 12.0 & 29.4 \\
$T = 1.0$ & 17.6 & 18.0 & 29.6 & 33.2 & 35.4 & 26.4 \\
Repetition penalty & 22.8 & 25.0 & 27.0 & 43.1 & 48.9 & 20.6 \\
Budget forcing & 33.4 & 34.2 & 10.9 & 53.7 & 59.0 & 9.6 \\
Loop stop & 33.6 & 34.1 & 10.9 & 53.0 & 58.2 & 7.9 \\
Thinking off, 8k & 43.3 & 67.7 & 3.8 & 56.7 & 95.6 & 1.2 \\
End-of-thinking bias & 39.6 & 47.5 & 18.8 & 58.2 & 77.6 & 9.1 \\
Thinking off, full budget & 48.6 & 85.9 & 8.3 & 58.1 & 98.7 & 1.5 \\
Textbook edit & 27.4 & 28.7 & 28.0 & 45.9 & 53.2 & 22.5 \\
Agent's edit & \textbf{51.7} & 69.1 & 21.9 & \textbf{67.6} & 93.1 & 11.9 \\
\midrule
Base, 81{,}920 tokens & 9.6 & 9.4 & 74.9 & -- & -- & -- \\
\bottomrule
\end{tabular}
\caption{Qwen3.5-0.8B on the test sets (4 samples per problem; 81{,}920-token row: first sample). Acc: accuracy,
authors' scoring (\%); Fin: answers finished (\%; a forced answer is cut at its 4{,}096-token cap); Tok: mean generated tokens (thousands).}
\label{tab:main}
\end{table}

%% file: tables/rules.tex
\begin{table*}[t]
\centering
\footnotesize
\setlength{\tabcolsep}{2.6pt}
\begin{tabular}{@{}lllllcc@{}}
\toprule
& & \multicolumn{3}{c}{Paired contrast: share rows in points, Own 32 rows in problems} & Reading$^{r}$ & Flagged$^{f}$ \\
\cmidrule(lr){3-5}
& Reach & Distilled $-$ own-answer & Edit $-$ own-answer & Edit $-$ distilled & edit\,/\,distilled & d/r/e \\
\midrule
MATH-500 & Full (32) & $+$0.7 [$-$0.4, 1.8], 2/18 & $+$0.2 [$-$0.8, 1.4] & $-$0.5 [$-$1.7, 0.8] & -- & 0/0/no \\
 & $k = 16$ & $+$4.2 [2.6, 5.7], 18/18 & $+$1.1 [$-$0.4, 2.6] & $-$3.1 [$-$4.8, $-$1.3] & -- & 0/0/no \\
 & $k = 4$ & $+$6.4 [4.7, 8.1], 18/18 & $+$1.7 [$-$0.2, 3.6] & $-$4.7 [$-$6.7, $-$2.7] & -- & 0/0/no \\
 & Own 32$^{e}$ & $+$19 [12, 27], 18/18 & $+$6 [$-$2, 15] & $-$13 [$-$22, $-$4] & below\,/\,$\le$4 & -- \\
\midrule
GSM8K & Full (32) & $+$1.5 [0.9, 2.2], 18/18 & $-$0.5 [$-$0.9, $-$0.2] & $-$2.0 [$-$2.8, $-$1.4] & -- & 3/0/no \\
 & $k = 16$ & $+$2.9 [2.3, 3.7], 18/18 & $-$0.6 [$-$1.1, $-$0.03] & $-$3.5 [$-$4.3, $-$2.7] & -- & 3/0/no \\
 & $k = 4$ & $+$5.6 [4.8, 6.5], 18/18 & $-$0.5 [$-$1.4, 0.4] & $-$6.1 [$-$7.3, $-$5.1] & -- & 3/2/no \\
 & Own 32$^{e}$ & $+$23 [16, 31], 18/18 & $-$9 [$-$16, $-$2] & $-$31 [$-$42, $-$21] & below\,/\,adds & -- \\
\midrule
MMLU-Pro math$^{c}$ & Full (32) & $+$6.0 [2.3, 9.7], 9/18 & $+$4.4 [0.6, 8.6] & $-$1.5 [$-$6.0, 2.9] & -- & 0/0/no \\
 & Own 32$^{e}$ & $-$7 [$-$17, 3], 0/18 & $-$20 [$-$32, $-$8] & $-$13 [$-$26, $-$0.01] & $\le$8\,/\,$\le$23 & -- \\
 & Own 32$^{e,b}$ & $+$17 [9, 26], 16/18 & $+$12 [4, 21] & $-$5 [$-$15, 5] & $\le$10\,/\,adds & -- \\
\bottomrule
\end{tabular}
\caption{The beyond-reach test on every set and decoding reach (95\% paired intervals). Distilled runs pooled minus own-answer runs pooled (bias answers), with the pairs above zero; the edit minus each pool. $^{r}$Against the decoding-only references and the own-answer runs (\S\ref{sec:method}; Table~\ref{tab:refs}), for the edit and for the distilled runs pooled: \emph{adds}, \emph{below}, or bound, $\le$ the largest addition not excluded (problems); the share rows have no decoding-only reference. $^{f}$Flagged by the main flag: distilled runs\,/\,own-answer runs\,/\,the edit. $^{c}$Chance-corrected. $^{e}$Each model with 32 samples of its own: problems per run, of those beyond the decoding reach (MMLU-Pro math: beyond chance). $^{b}$Per question, at least the right answers a binomial test on the model's own finished samples asks for.}
\label{tab:rules}
\end{table*}

%% file: tables/ctrl.tex
\begin{table}[t]
\centering
\footnotesize
\setlength{\tabcolsep}{2.4pt}
\begin{tabular}{@{}lccc@{}}
\toprule
& MATH-500 & GSM8K & MMLU-Pro \\
\midrule
\multicolumn{4}{@{}l}{\emph{Per-sample strict accuracy (\%)}} \\
\hspace{0.5em}Agent's edit & 43.8 & 57.2 & 43.5 \\
\hspace{0.5em}Own-answer (6) & 37.4--40.6 & 54.6--57.2 & 24.7--28.9 \\
\hspace{0.5em}Distilled (3) & 46.8--48.1 & 66.5--67.3 & 39.5--40.2 \\
\multicolumn{4}{@{}l}{\emph{Answers finished (\%)}} \\
\hspace{0.5em}Agent's edit & 64.1 & 91.0 & 78.7 \\
\hspace{0.5em}Own-answer (6) & 48.4--53.2 & 77.4--81.7 & 35.1--43.6 \\
\hspace{0.5em}Distilled (3) & 64.0--66.4 & 95.7--97.4 & 61.5--63.7 \\
\multicolumn{4}{@{}l}{\emph{Problems beyond reach}} \\
\hspace{0.5em}Agent's edit & 14 & 13 & 38 \\
\hspace{0.5em}Own-answer (6) & 7--13 & 15--23 & 9--20 \\
\hspace{0.5em}Distilled (3) & 15--20 & 37--42 & 33--38 \\
\multicolumn{4}{@{}l}{\emph{Own 32: problems solved beyond the reach$^{o}$}} \\
\hspace{0.5em}Agent's edit & 36 & 41 & $-$27 \\
\hspace{0.5em}Own-answer (6) & 28--31 & 44--57 & $-$12 to $-$4 \\
\hspace{0.5em}Distilled (3) & 44--52 & 72--73 & $-$19 to $-$5 \\
\hspace{0.5em}Sampled again$^{u}$ & 24 & 28 & $-$18 \\
\hspace{0.5em}Thinking off$^{z}$ & 55 & 59 & $-$22 \\
\bottomrule
\end{tabular}
\caption{The edit against the controls where the base model always loops (MATH-500, GSM8K, MMLU-Pro math: 426, 934 and 842 problems; 4 samples each). $^{o}$Of the problems beyond the decoding reach, those some finished sample of a model's own 32 solves; MMLU-Pro math chance-corrected (per run: Tables~\ref{tab:pass32} and~\ref{tab:own32runs}). $^{u}$The unedited model, four new samples under each of the eight fixes. $^{z}$The unedited model, 32 further samples with thinking off at the full budget.}
\label{tab:ctrl}
\end{table}

%% file: tables/second.tex
\begin{table}[t]
\centering
\small
\setlength{\tabcolsep}{4pt}
\begin{tabular}{@{}lrrrr@{}}
\toprule
& \multicolumn{2}{c}{MATH-500} & \multicolumn{2}{c}{GSM8K} \\
\cmidrule(lr){2-3}\cmidrule(l){4-5}
Qwen3.5-2B & Answers & Coverage & Answers & Coverage \\
\midrule
Textbook edit & 40 & 100.0 & 167 & 100.0 \\
Qwen3.5-4B & 326 & 87.7$^{f}$ & 399 & 94.2$^{f}$ \\
Fixes' own & -- & 89.0--100 & -- & 98.0--100 \\
\bottomrule
\end{tabular}
\caption{The flag on Qwen3.5-2B's always-looping problems, against the 2B's eight fixes: correct answers and the coverage (\%; $^{f}$flagged); fixes' own: each against the other seven.}
\label{tab:second}
\end{table}

%% file: tables/summary.tex
\begin{table*}[t]
\centering
\footnotesize
\setlength{\tabcolsep}{3.2pt}
\begin{tabular}{@{}lllll@{}}
\toprule
& MATH-500 & GSM8K & MMLU-Pro math & MedQA \\
\midrule
Accuracy, base $\to$ edit (\%) & 7.6 $\to$ 51.7 & 13.7 $\to$ 67.6 & 40.7 $\to$ 62.4 & 42.5 $\to$ 46.1 \\
Gain recovered$^{t}$, authors' / strict (\%) & 93 / 93 & 82 / 83 & 56 / 77 & 56 / 81 \\
\midrule
\multicolumn{5}{@{}l}{\emph{Whole test sets: what coverage cannot see}} \\
Edit $-$ best fix, strict, per sample & $+$3.1 [0.6, 5.6] & $+$9.3 [7.9, 10.8] & $+$8.6 [7.1, 10.1] & $+$1.7 [0.1, 3.3] \\
Both finish: edit / thinking off (\%) & 77.5 / 65.2 & 72.9 / 60.4 & 72.6 / 60.5 & -- \\
Edit $-$ voting thinking off, $\le$ edit's tokens & $+$0.1 [$-$2.6, 2.9] & $+$0.3 [$-$1.6, 2.2] & $+$9.4 [7.9, 10.9] & -- \\
\midrule
\multicolumn{5}{@{}l}{\emph{Beyond reach, on always-looping problems: correct answers within the decoding reach (\%)}} \\
Edit, all fixes / thinking on & 97.7 / 88.0 & 99.3 / 97.0 & 87.5$^{c}$ & -- \\
Qwen3.5-2B, random / most reliable & 92.0 / 93.8 & 96.6 / 97.1 & 52.9$^{c}$ & -- \\
The distilled 0.8B, three runs & 97.2--97.4 & 97.1--97.3 & 83.3--87.5$^{c}$ & -- \\
Own-answer fine-tunes, six runs & 97.5--98.7 & 98.3--98.9 & 88.4--94.4$^{c}$ & -- \\
Self-training, thinking-off answers, two runs & 95.1--96.4 & 98.5--98.8 & 82.8--82.9$^{c}$ & -- \\
\midrule
Released $-$ own (points) & $-$0.7 [$-$5.1, 2.9] & $-$1.7 [$-$3.3, $-$0.2] & $-$5.5 [$-$7.8, $-$3.3] & $-$0.7 [$-$4.0, 2.6] \\
Thinking off: edit $-$ base (points) & $+$3.7 [1.0, 6.5] & $-$2.0 [$-$3.7, $-$0.3] & $+$3.7 [1.8, 5.6] & -- \\
\quad same, where both finish$^{f}$ & $-$0.6 [$-$3.6, 2.5] & $-$2.5 [$-$4.3, $-$0.7] & $-$0.6 [$-$2.6, 1.5] & -- \\
\bottomrule
\end{tabular}
\caption{Table~\ref{tab:headline} in detail, with MedQA (Qwen3.5-0.8B, 4 samples per problem; points unless \%; 95\% intervals). $^{t}$Thinking off, full budget; MedQA: $T = 1.0$. $^{c}$Chance-corrected. Most reliable: the 2B's answers on as many problems as the edit solves, those it solves most often. $^{f}$On the samples both finish.
}
\label{tab:summary}
\end{table*}

%% file: tables/pass32runs.tex
\begin{table}[t]
\centering
\footnotesize
\setlength{\tabcolsep}{2pt}
\begin{tabular}{@{}lrrrrl@{}}
\toprule
& & \multicolumn{2}{c}{Beyond} & & \\
\cmidrule(lr){3-4}
Run & Solved & 4 & 32 & Share & Edit $-$ run \\
\midrule
Agent's edit & 336 & 14 & 36 & 1.9 [1.1, 3.0] &  \\
\multicolumn{6}{@{}l}{\emph{Own-answer, bias answers}} \\
seed 0 & 327 & 12 & 30 & 1.9 [1.0, 3.0] & $+$6 [$-$3, 15] \\
seed 1 & 339 & 10 & 31 & 1.8 [1.0, 2.8] & $+$5 [$-$4, 14] \\
seed 2 & 333 & 13 & 31 & 1.8 [1.0, 2.7] & $+$5 [$-$4, 14] \\
seed 3 & 329 & 13 & 28 & 1.7 [0.9, 2.6] & $+$8 [$-$3, 19] \\
aligned 1 & 325 & 11 & 28 & 1.8 [0.9, 2.8] & $+$8 [$-$2, 19] \\
aligned 2 & 334 & 7 & 31 & 1.7 [0.9, 2.6] & $+$5 [$-$5, 15] \\
\multicolumn{6}{@{}l}{\emph{Distilled from Qwen3.5-9B}} \\
seed 0 & 366 & 16 & 51 & 2.9 [1.9, 4.2] & $-$15 [$-$25, $-$5] \\
seed 1 & 355 & 15 & 44 & 2.6 [1.6, 3.8] & $-$8 [$-$19, 3] \\
seed 2 & 367 & 20 & 52 & 2.7 [1.8, 3.8] & $-$16 [$-$26, $-$6] \\
\bottomrule
\end{tabular}
\caption{Each model's own 32 samples on MATH-500's 426 always-looping problems: problems solved, those of the 101 beyond the decoding reach (with the first 4 and all 32 samples), beyond-reach answers as a share of the model's correct answers (\%), and the edit minus each run in problems (95\% paired intervals).}
\label{tab:pass32}
\end{table}

%% file: tables/acc12.tex
\begin{table*}[t]
\centering
\footnotesize
\setlength{\tabcolsep}{3pt}
\begin{tabular}{@{}llll@{}}
\toprule
& MATH-500 & GSM8K & MMLU-Pro math \\
\midrule
\multicolumn{4}{@{}l}{\emph{Thinking off at the full budget: accuracy, points}} \\
Edit $-$ base & $+$3.7 [1.0, 6.5] & $-$2.0 [$-$3.7, $-$0.3] & $+$3.7 [1.8, 5.6] \\
Self-training $-$ base, two runs averaged & $+$0.9 [$-$1.4, 3.2] & $+$2.0 [0.3, 3.7] & $+$1.9 [0.2, 3.5] \\
Distilled $-$ own-answer, pooled; pairs & $+$3.7 [1.8, 5.6]; 13/18 & $+$3.6 [2.3, 4.9]; 16/18 & $+$8.5 [7.2, 9.8]; 18/18 \\
Edit $-$ own-answer, pooled & $+$2.5 [$-$0.05, 5.0] & $-$8.6 [$-$10.2, $-$7.0] & $+$7.5 [5.7, 9.3] \\
Edit $-$ distilled, pooled & $-$1.2 [$-$3.6, 1.3] & $-$12.3 [$-$14.0, $-$10.6] & $-$1.0 [$-$2.7, 0.8] \\
Reading, pair\,/\,pooled & --\,/\,bound & --\,/\,none & adds\,/\,adds \\
Own-answer (six) $-$ base, pooled & $+$1.2 [$-$1.2, 3.6] & $+$6.6 [5.1, 8.1] & $-$3.8 [$-$5.4, $-$2.2] \\
Distilled (three) $-$ base, pooled & $+$4.9 [2.5, 7.2] & $+$10.2 [8.6, 11.9] & $+$4.7 [3.1, 6.3] \\
Cut (\%)$^{k}$ & 0\,/\,10--12\,/\,33--41\,/\,21--22\,/\,17 & 0\,/\,1\,/\,6--9\,/\,1--2\,/\,2 & 0\,/\,11\,/\,40--49\,/\,19\,/\,16 \\
\multicolumn{4}{@{}l}{\emph{Thinking off, on the samples both finish: accuracy, points}} \\
Edit $-$ base & $-$0.6 [$-$3.6, 2.5] & $-$2.5 [$-$4.3, $-$0.7] & $-$0.6 [$-$2.6, 1.5] \\
Edit $-$ own-answer, pooled & $-$12.1 [$-$15.0, $-$9.2] & $-$11.2 [$-$12.9, $-$9.6] & $-$10.8 [$-$12.9, $-$8.8] \\
Edit $-$ distilled, pooled & $-$8.4 [$-$11.0, $-$5.8] & $-$12.7 [$-$14.4, $-$11.0] & $-$7.4 [$-$9.3, $-$5.6] \\
Self-training $-$ base & $-$0.8 [$-$3.6, 2.0] & $+$1.9 [0.1, 3.6] & $+$0.5 [$-$1.5, 2.4] \\
\midrule
\multicolumn{4}{@{}l}{\emph{End-of-thinking bias ($+8$): accuracy, points}} \\
Own-answer (six, averaged) $-$ base & $-$27.4 [$-$30.7, $-$24.2] & $-$38.9 [$-$41.2, $-$36.8] & $-$25.8 [$-$27.9, $-$23.6] \\
Cut (\%): own-answer\,/\,base & 96--98\,/\,61 & 87--91\,/\,30 & 97--99\,/\,56 \\
\bottomrule
\end{tabular}
\caption{The accuracy criterion applied to the edit and the controls on the problems the test uses (4 samples; finished and right; 95\% paired intervals). Reading as in Table~\ref{tab:rules}, with accuracy as the measure; --, the controls do not separate. Under the bias the own-answer runs' cut samples end repeating the end-of-thinking token. Two different contrasts happen to read $+$3.7 [1.8, 5.6]. $^{k}$Edit\,/\,self-training\,/\,own-answer\,/\,distilled\,/\,base.}
\label{tab:acc12}
\end{table*}

%% file: tables/own32runs.tex
\begin{table*}[t]
\centering
\footnotesize
\setlength{\tabcolsep}{6pt}
\begin{tabular}{@{}lccccc@{}}
\toprule
& \multicolumn{2}{c}{GSM8K} & \multicolumn{3}{c}{MMLU-Pro math} \\
\cmidrule(lr){2-3}\cmidrule(lr){4-6}
Run & solved & matched$^{m}$ & beyond chance & test$^{b}$ & matched$^{m}$ \\
\midrule
Agent's edit & 41 & 26 & $-$27 & 25 & $-$13 \\
Distilled s0 & 72 & 58 & $-$5 & 28 & 5 \\
Distilled s1 & 72 & 54 & $-$17 & 27 & $-$3 \\
Distilled s2 & 73 & 57 & $-$19 & 34 & 1 \\
Own-answer s0 & 44 & 38 & $-$5 & 13 & $-$3 \\
Own-answer s1 & 46 & 38 & $-$6 & 17 & $-$9 \\
Own-answer s2 & 51 & 43 & $-$7 & 16 & $-$5 \\
Own-answer s3 & 46 & 41 & $-$12 & 11 & $-$8 \\
Data-aligned s1 & 53 & 46 & $-$8 & 11 & $-$7 \\
Data-aligned s2 & 57 & 46 & $-$4 & 10 & 0 \\
\midrule
Unedited, sampled again$^{u}$ & 28 & -- & $-$18 & 5 & -- \\
Unedited, thinking off$^{z}$ & 59 & -- & $-$22 & 13 & -- \\
\bottomrule
\end{tabular}
\caption{Each model's own 32 samples on the problems beyond the decoding reach (GSM8K 101, MMLU-Pro math 185; MATH-500: Table~\ref{tab:pass32}). $^{m}$Each model's samples kept up to as many finished ones as the unedited model sampled again finishes on the problem. $^{b}$The binomial test on the model's own finished samples. $^{u,z}$As in Table~\ref{tab:ctrl}.}
\label{tab:own32runs}
\end{table*}

%% file: tables/reg.tex
\begin{table*}[t]
\centering
\footnotesize
\setlength{\tabcolsep}{4pt}
\begin{tabular}{@{}p{0.50\textwidth}lp{0.31\textwidth}@{}}
\toprule
Rule registered & Written (PST) & Known then \\
\midrule
Two more distilled runs and two data-aligned own-answer runs, each read run by run & Nov 10, 07:05 & one distilled and four own-answer runs under the full decoding reach; first committed with the new runs' first results \\
Self-training with thinking off adds if its averaged difference from the base model lies above zero on two of three sets & Nov 10, 17:57 & its shares beyond reach, not its thinking-off accuracy \\
MATH-500 at 32 samples per model: the 18 pairs and the edit against each pool & Nov 11, 03:59 & one distilled and one own-answer run and the edit at 32 samples \\
One rule for every set (every pair primary; none, adds or bound with half the distilled advantage; strict counting); 32 samples on GSM8K and MMLU-Pro math; the accuracy checks & Nov 11, 21:44 & every MATH-500 result at 32 samples; every share under the decoding reaches \\
MMLU-Pro math: the count beyond chance stays primary, the binomial test beside it & Nov 11, 22:03 & the models' finishing rates there, not their correctness \\
The controls with thinking off, read by the same rule & Nov 12, 02:38 & GSM8K at 32 samples; the edit with thinking off against the base model \\
The unedited model sampled again as a reference, unmatched and matched by finished samples; accuracy where both finish, with thinking off & Nov 14, 15:28 & every result above; finished shares with thinking off; the solved counts of the unedited model sampled again \\
\bottomrule
\end{tabular}
\caption{Registered rules, when each was written down (2026) and what was known then. The dated record, with each rule's commit and push times, is released with the code.}
\label{tab:reg}
\end{table*}

%% file: tables/refs.tex
\begin{table*}[t]
\centering
\footnotesize
\setlength{\tabcolsep}{3pt}
\begin{tabular}{@{}llll@{}}
\toprule
Model $-$ reference & MATH-500 & GSM8K & MMLU-Pro math \\
\midrule
\multicolumn{4}{@{}l}{\emph{The unedited model sampled again, solving 24, 28 and $-$18$^{c}$}} \\
Edit & $+$12 [2, 22] ($+$1 [$-$9, 11]) & $+$13 [4, 23] ($-$2 [$-$10, 6]) & $-$8 [$-$24, 6] ($+$5 [$-$11, 21]) \\
Edit, cut samples too & $+$14 [4, 24] ($+$2 [$-$8, 12]) & $+$13 [4, 23] ($-$2 [$-$10, 6]) & $+$11 [$-$2, 24] ($+$17 [2, 33]) \\
Edit, fixed-$k$ count & -- & -- & $+$24 [15, 34] ($+$8 [3, 14]) \\
Edit, binomial test & -- & -- & $+$20 [9, 31] ($+$3 [$-$4, 10]) \\
Own-answer, pooled & $+$6 [$-$2, 14] ($+$3 [$-$5, 11]) & $+$22 [14, 30] ($+$14 [7, 22]) & $+$12 [$-$1, 24] ($+$13 [0.3, 25]) \\
Distilled, pooled & $+$25 [16, 35] ($+$13 [4, 21]) & $+$44 [32, 57] ($+$28 [19, 39]) & $+$5 [$-$9, 18] ($+$20 [7, 32]) \\
\midrule
\multicolumn{4}{@{}l}{\emph{The unedited model with thinking off, solving 55, 59 and $-$22$^{c}$}} \\
Edit & $-$19 [$-$31, $-$8] ($-$21 [$-$33, $-$9]) & $-$18 [$-$30, $-$7] ($-$18 [$-$30, $-$7]) & $-$4 [$-$18, 10] ($-$6 [$-$20, 8]) \\
Edit, cut samples too & $-$17 [$-$28, $-$6] ($-$19 [$-$31, $-$8]) & $-$18 [$-$30, $-$7] ($-$18 [$-$30, $-$7]) & $-$9 [$-$21, 3] ($-$10 [$-$23, 2]) \\
Edit, fixed-$k$ count & -- & -- & $+$10 [0, 21] ($+$3 [$-$6, 13]) \\
Edit, binomial test & -- & -- & $+$12 [1, 23] ($+$8 [$-$3, 18]) \\
Own-answer, pooled & $-$25 [$-$36, $-$15] ($-$25 [$-$36, $-$15]) & $-$10 [$-$19, 1] ($-$10 [$-$19, 1]) & $+$16 [4, 28] ($+$16 [4, 28]) \\
Distilled, pooled & $-$6 [$-$16, 4] ($-$6 [$-$16, 4]) & $+$13 [4, 24] ($+$13 [3, 23]) & $+$9 [$-$5, 23] ($+$9 [$-$5, 23]) \\
\bottomrule
\end{tabular}
\caption{The two references on the problems beyond the decoding reach (101, 101 and 185): a model's problems solved minus the reference's, with 32 samples each (95\% paired intervals); in parentheses, each model's samples kept up to as many finished ones as the reference finishes on the problem. Strict counting, MMLU-Pro math beyond chance$^{c}$, unless marked. $^{c}$Chance-corrected.}
\label{tab:refs}
\end{table*}

%% file: tables/med.tex
\begin{table}[t]
\centering
\small
\setlength{\tabcolsep}{2.2pt}
\begin{tabular}{@{}lrrrrrr@{}}
\toprule
& \multicolumn{3}{c}{MedQA} & \multicolumn{3}{c}{MMLU-Pro h.} \\
\cmidrule(lr){2-4}\cmidrule(l){5-7}
& Strict & Acc & Fin & Strict & Acc & Fin \\
\midrule
Base & 37.4 & 42.5 & 76.7 & 31.1 & 33.9 & 84.5 \\
Prompt only & 28.6 & 38.7 & 55.7 & -- & -- & -- \\
Thinking off, 8k & 39.5 & 40.0 & 96.9 & 29.9 & 30.3 & 97.8 \\
$T = 1.0$ & 44.4 & 44.5 & 99.3 & 34.5 & 34.6 & 99.4 \\
Budget forcing & 42.2 & 44.5 & 87.5 & 33.6 & 35.0 & 92.1 \\
Loop stop & 42.0 & 44.4 & 87.2 & 33.8 & 35.0 & 92.2 \\
Medical-direction edit & 43.6 & 44.9 & 94.5 & 35.1 & \textbf{35.7} & 95.6 \\
Agent's edit & \textbf{46.0} & \textbf{46.1} & 99.9 & \textbf{35.4} & 35.5 & 99.9 \\
\bottomrule
\end{tabular}
\caption{Qwen3.5-0.8B on medical QA (4 samples per question; \%). Strict counts cut answers as wrong. The agent's edit
is transferred unchanged from mathematics.}
\label{tab:med}
\end{table}

%% file: tables/pair.tex
\begin{table*}[t]
\centering
\footnotesize
\setlength{\tabcolsep}{3.0pt}
\begin{tabular}{@{}llll@{}}
\toprule
& MATH-500 & GSM8K & MMLU-Pro math \\
\midrule
Agent's edit: released / own (\%) & 97.8 / 98.5 & 94.2 / 95.9 & 84.6 / 90.1 \\
\multicolumn{4}{@{}l}{\emph{Released $-$ own, same problems (points, 95\% CI)}} \\
Agent's edit & $-$0.7 [$-$5.1, 2.9] & $-$1.7 [$-$3.3, $-$0.2] & $-$5.5 [$-$7.8, $-$3.3] \\
$T = 1.0$ & $+$1.9 [0, 4.7] & $-$0.2 [$-$1.7, 1.5] & $-$4.2 [$-$6.8, $-$1.6] \\
Budget forcing & $+$0.8 [$-$1.5, 3.8] & $+$0.4 [$-$1.2, 2.0] & $-$1.9 [$-$3.7, $-$0.2] \\
Loop stop & $-$1.2 [$-$5.7, 2.2] & $-$0.2 [$-$1.7, 1.3] & $-$1.3 [$-$3.1, 0.4] \\
Textbook edit & $-$0.5 [$-$4.7, 3.1] & $-$1.4 [$-$3.4, 0.5] & -- \\
Prompt only & $+$2.9 [0, 7.4] & $+$0.9 [0, 2.3] & -- \\
End-of-thinking bias & $-$2.2 [$-$5.4, 0.7] & $-$5.7 [$-$8.1, $-$3.3] & -- \\
Repetition penalty & $-$5.3 [$-$10.9, $-$0.8] & $-$2.7 [$-$4.8, $-$0.6] & -- \\
Thinking off, full budget & $-$9.3 [$-$15.7, $-$3.7] & $-$13.9 [$-$17.1, $-$10.9] & $-$17.7 [$-$21.6, $-$14.0] \\
Thinking off, 8k & $-$9.1 [$-$15.2, $-$3.7] & $-$15.1 [$-$18.4, $-$12.0] & $-$18.9 [$-$22.8, $-$15.1] \\
R1-Distill: textbook edit $-$ $T{=}0.6$ & $+$9.1 [0, 20.0] & -- & -- \\
Problems paired & 34--68 & 175--366 & 294--342 \\
\midrule
& MedQA & MMLU-Pro health & \\
\midrule
Agent's edit: released / own (\%) & 42.2 / 42.9 & 33.4 / 34.4 & \\
\multicolumn{4}{@{}l}{\emph{Released $-$ own, same problems (points, 95\% CI)}} \\
Agent's edit & $-$0.7 [$-$4.0, 2.6] & $-$1.1 [$-$5.3, 3.2] & \\
$T = 1.0$ & $-$2.3 [$-$5.8, 1.1] & $-$3.4 [$-$7.6, 1.0] & \\
Budget forcing & $-$1.3 [$-$5.6, 2.9] & $-$3.0 [$-$8.7, 2.6] & \\
Loop stop & $-$0.6 [$-$5.1, 3.9] & $-$2.9 [$-$9.2, 3.2] & \\
Medical-direction edit & $-$2.0 [$-$5.7, 1.6] & $+$2.1 [$-$1.0, 5.3]$^{*}$ & \\
Prompt only & $+$1.1 [$-$3.9, 6.0] & -- & \\
Thinking off, 8k & $-$7.9 [$-$11.8, $-$4.1] & $-$6.6 [$-$11.4, $-$1.8] & \\
Problems paired & 324--630 & 198--413 & \\
\bottomrule
\end{tabular}
\caption{Released answers against the model's own on the same problems (Qwen3.5-0.8B; problems on which the base model
finishes some samples). R1-Distill: its textbook edit's released answers against $T=0.6$ on the same 55 problems
(81.8\% against 72.7\%). $^{*}$Eight samples per question.}
\label{tab:pair}
\end{table*}

%% file: tables/majk.tex
\begin{table*}[t]
\centering
\small
\setlength{\tabcolsep}{6pt}
\begin{tabular}{@{}lrrrr@{}}
\toprule
& maj@1 & maj@2 & maj@3 & maj@4 \\
\midrule
\multicolumn{5}{@{}l}{\emph{MATH-500}} \\
Agent's edit & 51.7 (21.9) & 59.1 (43.7) & 62.7 (65.6) & 64.8 (87.4) \\
Thinking off, full budget & 48.6 (8.3) & 51.5 (16.6) & 56.4 (24.9) & 60.1 (33.2) \\
Thinking off, 8k & 43.3 (3.8) & 47.7 (7.5) & 52.7 (11.3) & 56.1 (15.1) \\
\midrule
\multicolumn{5}{@{}l}{\emph{GSM8K}} \\
Agent's edit & 67.6 (11.9) & 71.3 (23.7) & 74.2 (35.6) & 76.0 (47.4) \\
Thinking off, full budget & 58.1 (1.5) & 58.3 (3.1) & 64.2 (4.6) & 67.2 (6.1) \\
Thinking off, 8k & 56.7 (1.2) & 57.6 (2.4) & 63.6 (3.6) & 66.6 (4.8) \\
\midrule
\multicolumn{5}{@{}l}{\emph{MMLU-Pro math}} \\
Agent's edit & 62.4 (16.9) & 62.5 (33.9) & 65.4 (50.8) & 66.5 (67.7) \\
Thinking off, full budget & 52.9 (6.5) & 52.9 (13.1) & 57.8 (19.6) & 59.8 (26.2) \\
Thinking off, 8k & 49.9 (3.4) & 50.0 (6.8) & 54.4 (10.1) & 56.9 (13.5) \\
\bottomrule
\end{tabular}
\caption{Majority vote over $j$ of the four samples (accuracy \%, authors' scoring; in parentheses, generated tokens per problem, thousands). A sample without an answer does not vote (a subset with no vote is wrong); ties are split evenly.}
\label{tab:majk}
\end{table*}

%% file: tables/kshare.tex
\begin{table*}[t]
\centering
\small
\setlength{\tabcolsep}{4pt}
\begin{tabular}{@{}lrrrrrrrrrr@{}}
\toprule
& \multicolumn{6}{c}{MATH-500} & \multicolumn{4}{c}{GSM8K} \\
\cmidrule(lr){2-7}\cmidrule(l){8-11}
& Full & On & L4--5 & $k{=}4$ & $k{=}16$ & $k{=}57$ & Full & On & $k{=}4$ & $k{=}16$ \\
\midrule
Agent's edit & 2.3 & 12.0 & 4.2 & 30.4 & 9.5 & 2.1 & 0.7 & 3.0 & 16.1 & 2.8 \\
Bias answers, seed 0 & 2.0 & 10.0 & 3.4 & 28.2 & 8.1 & 1.8 & 1.1 & 4.2 & 16.4 & 3.2 \\
Bias answers, seed 1 & 2.0 & 10.7 & 3.4 & 29.4 & 8.5 & 1.7 & 1.3 & 5.4 & 17.6 & 3.7 \\
Bias answers, seed 2 & 2.1 & 10.9 & 3.1 & 28.1 & 8.4 & 1.8 & 1.7 & 4.4 & 16.9 & 3.9 \\
Bias answers, seed 3 & 2.4 & 9.7 & 4.0 & 28.7 & 8.5 & 2.4 & 1.1 & 4.3 & 16.2 & 3.2 \\
Bias answers, data-aligned, seed 1 & 2.5 & 9.8 & 5.2 & 28.5 & 8.6 & 2.0 & 1.2 & 4.7 & 15.7 & 3.1 \\
Bias answers, data-aligned, seed 2 & 1.3 & 10.3 & 2.0 & 28.7 & 8.7 & 1.1 & 1.3 & 5.1 & 16.4 & 3.4 \\
Self-training, thinking-off answers, seed 1 & 3.6 & 20.4 & 6.4 & 35.7 & 13.0 & 3.0 & 1.2 & 5.6 & 16.3 & 3.3 \\
Self-training, thinking-off answers, seed 2 & 4.9 & 21.3 & 7.9 & 36.7 & 14.3 & 4.6 & 1.5 & 6.1 & 16.9 & 3.6 \\
Distilled 0.8B, seed 0 & 2.8 & 16.3 & 5.1 & 33.7 & 11.7 & 2.6 & 2.9 & 8.9 & 22.8 & 6.8 \\
Distilled 0.8B, seed 1 & 2.6 & 17.1 & 2.6 & 35.7 & 13.0 & 2.5 & 2.7 & 7.7 & 21.5 & 5.9 \\
Distilled 0.8B, seed 2 & 2.8 & 18.4 & 3.7 & 35.6 & 13.2 & 2.7 & 2.8 & 8.5 & 22.3 & 6.3 \\
Qwen3.5-2B & 8.0 & 25.7 & 12.5 & 44.3 & 21.1 & 7.7 & 3.4 & 8.6 & 23.1 & 7.1 \\
Qwen3.5-4B & 22.0 & 43.6 & 31.9 & 58.6 & 37.2 & 21.5 & 8.9 & 18.5 & 34.0 & 14.7 \\
Qwen3.5-9B & 22.6 & 44.3 & 32.6 & 59.2 & 37.9 & 22.1 & 9.2 & 19.0 & 34.4 & 15.1 \\
\bottomrule
\end{tabular}
\caption{Share of correct answers beyond the unedited 0.8B's reach on its always-looping problems (\%): against the decoding reach (32 samples), its six thinking-on fixes (On), MATH levels 4--5 only, and $k$ samples (MATH-500: from the reach samples and seven held-out decodings, 57 samples; GSM8K: from the reach samples).}
\label{tab:k}
\end{table*}

%% file: tables/runs.tex
\begin{table}[t]
\centering
\footnotesize
\setlength{\tabcolsep}{1.7pt}
\begin{tabular}{@{}lrrrrrr@{}}
\toprule
& \multicolumn{2}{c}{MATH-500} & \multicolumn{2}{c}{GSM8K} & \multicolumn{2}{c}{MMLU-Pro$^{c}$} \\
\cmidrule(lr){2-3}\cmidrule(lr){4-5}\cmidrule(l){6-7}
& Corr. & Beyond & Corr. & Beyond & Corr. & Beyond \\
\midrule
Agent's edit & 747 & 2.3 & 2{,}138 & 0.7 & 1{,}465 & 12.5 \\
\multicolumn{7}{@{}l}{\emph{Own-answer, bias answers}} \\
\quad seed 0 & 660 & 2.0 & 2{,}061 & 1.1 & 831 & 8.7 \\
\quad seed 1 & 691 & 2.0 & 2{,}127 & 1.3 & 975 & 7.7 \\
\quad seed 2 & 667 & 2.1 & 2{,}137 & 1.7 & 910 & 8.3 \\
\quad seed 3 & 679 & 2.4 & 2{,}062 & 1.1 & 886 & 5.6 \\
\quad aligned 1 & 651 & 2.5 & 2{,}041 & 1.2 & 890 & 6.2 \\
\quad aligned 2 & 638 & 1.3 & 2{,}092 & 1.3 & 900 & 11.6 \\
\multicolumn{7}{@{}l}{\emph{Self-training, thinking-off answers}} \\
\quad seed 1 & 731 & 3.6 & 1{,}879 & 1.2 & 1{,}315 & 17.1 \\
\quad seed 2 & 738 & 4.9 & 1{,}913 & 1.5 & 1{,}291 & 17.2$^{\ast}$ \\
\multicolumn{7}{@{}l}{\emph{Distilled from Qwen3.5-9B}} \\
\quad seed 0 & 798 & 2.8 & 2{,}515 & 2.9$^{\ast}$ & 1{,}342 & 16.7 \\
\quad seed 1 & 805 & 2.6 & 2{,}484 & 2.7$^{\ast}$ & 1{,}331 & 12.9 \\
\quad seed 2 & 819 & 2.8 & 2{,}483 & 2.8$^{\ast}$ & 1{,}353 & 12.5 \\
Qwen3.5-2B & 1{,}076 & 8.0$^{\ast}$ & 2{,}450 & 3.4$^{\ast}$ & 2{,}788 & 47.1$^{\ast}$ \\
\bottomrule
\end{tabular}
\caption{Every run, full decoding reach: correct answers and the share beyond reach (\%; $^{\ast}$flagged). $^{c}$Chance-corrected.}
\label{tab:runs}
\end{table}

%% file: tables/both.tex
\begin{table*}[t]
\centering
\small
\setlength{\tabcolsep}{6pt}
\begin{tabular}{@{}llll@{}}
\toprule
Edit $-$ fix & MATH-500 & GSM8K & MMLU-Pro math \\
\midrule
$T = 1.0$ & $+$0.0 [$-$1.8, 1.8] & $-$2.9 [$-$4.1, $-$1.6] & $-$0.5 [$-$1.8, 0.8] \\
Budget forcing & $-$1.1 [$-$2.9, 0.6] & $-$3.4 [$-$4.5, $-$2.3] & $-$1.4 [$-$2.5, $-$0.3] \\
Loop stop & $-$1.1 [$-$2.8, 0.5] & $-$3.1 [$-$4.2, $-$2.1] & $-$1.7 [$-$2.8, $-$0.7] \\
End-of-thinking bias & $+$5.1 [2.4, 7.8] & $+$4.0 [2.5, 5.4] & -- \\
Repetition penalty & $+$4.6 [2.2, 7.1] & $-$0.8 [$-$2.1, 0.4] & -- \\
Prompt only & $+$0.0 [0, 0] & $-$2.1 [$-$3.9, $-$0.2] & -- \\
Thinking off, 8k & $+$14.4 [11.4, 17.5] & $+$13.1 [11.4, 14.8] & $+$12.0 [10.3, 13.7] \\
Thinking off, full budget & $+$12.4 [9.4, 15.4] & $+$12.6 [10.9, 14.3] & $+$12.1 [10.4, 13.7] \\
\bottomrule
\end{tabular}
\caption{On the samples both the edit and a fix finish (same problem and index), the edit's accuracy minus the fix's (points, 95\% intervals over problems). On MMLU-Pro math three fixes ran on part of the set only.}
\label{tab:both}
\end{table*}

%% file: tables/mix.tex
\begin{table}[t]
\centering
\small
\setlength{\tabcolsep}{4pt}
\begin{tabular}{@{}lrrrr@{}}
\toprule
& \multicolumn{2}{c}{MATH-500} & \multicolumn{2}{c}{GSM8K} \\
\cmidrule(lr){2-3}\cmidrule(l){4-5}
Mixed into & 2B & Dist. & 2B & Dist. \\
\midrule
$T = 1.0$ & 35 & 60 & 20 & 25 \\
Budget forcing & 15 & 35 & 15 & 15 \\
Loop stop & 15 & 35 & 15 & 15 \\
End-of-thinking bias & 15 & $>$60 & 10 & 20 \\
Repetition penalty & 20 & 60 & 15 & 20 \\
Prompt only & 70 & $>$80 & 50 & 60 \\
Thinking off, 8k & 20 & $>$50 & 15 & 20 \\
Thinking off, full budget & 25 & $>$50 & 15 & 25 \\
\midrule
Agent's edit$^{a}$ & $>$50 & $>$50 & 50 & $>$50 \\
\bottomrule
\end{tabular}
\caption{Share (\%) of Qwen3.5-2B's or the distilled 0.8B's correct answers, mixed into a fix's, from which the mixture falls below the fix's own coverage against the other seven ($>$$x$: at no share tested, up to $x$). $^{a}$Into the edit's answers, judged by the flag rule.}
\label{tab:mix}
\end{table}

%% file: tables/mmlu.tex
\begin{table}[t]
\centering
\small
\setlength{\tabcolsep}{2.0pt}
\begin{tabular}{@{}lrrrrrr@{}}
\toprule
& \multicolumn{3}{c}{All 1{,}351} & \multicolumn{3}{c}{Never finished} \\
\cmidrule(lr){2-4}\cmidrule(l){5-7}
& Acc & Strict & Fin & Acc & Strict & Fin \\
\midrule
Base & 40.7 & 22.2 & 24.2 & 20.0 & 0.0 & 0.0 \\
$T = 1.0$ & 51.2 & 39.7 & 48.0 & 31.5 & 15.6 & 22.9 \\
Budget forcing & 50.8 & 40.7 & 44.6 & 30.5 & 16.9 & 19.3 \\
Loop stop & 50.8 & 41.6 & 45.5 & 30.1 & 17.4 & 19.6 \\
End-of-thinking bias & -- & -- & -- & 33.6 & 26.5 & 44.2 \\
Repetition penalty & -- & -- & -- & 31.7 & 21.2 & 30.1 \\
Prompt only & -- & -- & -- & 19.1 & 1.3 & 1.5 \\
Thinking off, 8k & 49.9 & 46.7 & 74.7 & 34.6 & 29.7 & 61.6 \\
Thinking off, full budget & 52.9 & 51.5 & 89.7 & 37.7 & 35.6 & 84.1 \\
Agent's edit & 62.4 & 60.1 & 86.4 & 47.0 & 43.5 & 78.7 \\
\bottomrule
\end{tabular}
\caption{Qwen3.5-0.8B on MMLU-Pro math (4 samples per question; \%), on all questions and on the 842 the base model never finishes, where all eight fixes ran. Acc: authors' scoring; Strict: cut answers wrong; Fin: finished.}
\label{tab:mmlu}
\end{table}